\documentclass[10pt,twocolumn,letterpaper]{article}

 \usepackage[pagenumbers]{cvpr} 

\definecolor{cvprblue}{rgb}{0.21,0.49,0.74}
\usepackage[pagebackref,breaklinks,colorlinks,allcolors=cvprblue]{hyperref}
\usepackage{upgreek}
\usepackage{mathtools}

\def\paperID{} 
\def\confName{CVPR}
\def\confYear{2026}

\title{PoseGravity: Pose Estimation from Points and Lines with Axis Prior}

\author{Akshay Chandrasekhar\\
BallerTV
}

\begin{document}
\maketitle
\begin{abstract}
    This paper presents a fast and accurate algorithm for estimating absolute camera pose given a known rotation axis. Existing methods are typically limited to line features, two-point cases, or local optimization routines. The proposed approach provides a complete solution for arbitrary combinations of point and line correspondences, producing the globally optimal solution in all cases. The special cases of minimal and planar configurations are highlighted for their practical closed-form solutions. As a theoretical consequence of this work, the general pose problem (PnPnL) is characterized as a function of only two variables. Extensive experiments demonstrate the efficiency and accuracy of the proposed method.
\end{abstract}
\section{Introduction} \label{sec:intro}
Estimating camera pose from image features and known 3D geometry is crucial for many computer vision applications such as augmented reality, visual odometry, and structure from motion. In certain scenarios, prior knowledge of the camera's rotation axis is available (\eg IMU measurement, detected vanishing point, or domain assumptions). Using this prior knowledge, the resulting problem reduces from six degrees of freedom (three for orientation, three for position) to four degrees of freedom as the rotation matrix only has one degree of freedom to rotate about the given axis. Some benefits of this simplified problem include reducing the minimum correspondences needed, improving the robustness of estimation to noise, and providing a quick initial solution for the general estimation problem. Previous solutions have been proposed that solve this problem for specific types of configurations or correspondences. In this paper, a fast and accurate algorithm is presented that generalizes the solution further by solving the problem for both point and line feature correspondences in minimal and overconstrained cases.

\begin{figure}
    \centering
    \includegraphics[scale=1.]{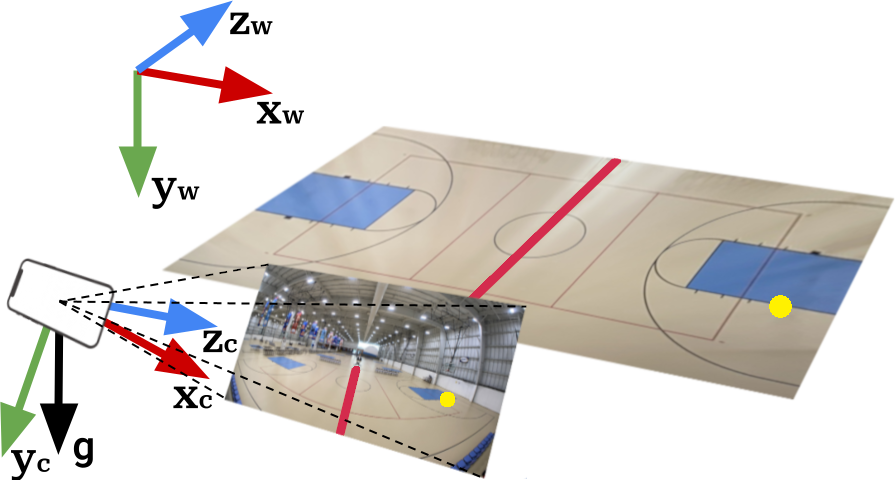}
    \caption{Example task illustration. A smartphone (coordinate frame $\mathbf{x_c},\mathbf{y_c},\mathbf{z_c}$) captures an image of a basketball court (coordinate frame $\mathbf{x_w},\mathbf{y_w},\mathbf{z_w}$) while measuring the gravity vector $\mathbf{g}$ with its IMU. Given image features matching known court locations (\eg red line, yellow point), we estimate the camera's position and orientation relative to the scene. The gravity vector, which observes the world y-axis in the camera's frame, provides prior information on the camera's orientation, greatly aiding estimation.}
    \label{fig:illustration}
\end{figure}

\section{Relevant Work}
\label{sec:relevant-work}

Estimating camera pose from point correspondences is a well-studied problem. The first known solution was proposed by Grunert \cite{Grunert} in 1841 for the minimal case of three projective point feature correspondences (P3P). Since then, numerous other analytical solutions have been proposed, typically framing the solutions as the roots of a quartic equation \cite{Ke}\cite{Kneip}. Persson \cite{LambdaTwist} took an alternative approach to the problem by formulating it as the intersection of two conic sections which involves finding a single root of a cubic equation instead.

Beyond the original P3P problem, solutions also emerged for the overconstrained pose estimation problem (PnP) for $n>3$ points. EPnP \cite{EPnP} was one of the earliest solutions to the general PnP problem that ran O($n$) time for \emph{n} $\ge$ 4. Terzakis \cite{SQPnP} provided a PnP solution by formulating the problem as an quadratic minimization problem followed by an iterative sequential quadratic program that was guaranteed to find the global minimum in a least squares sense.

Lines are another projective feature commonly used in the pose estimation problem. Analogous to the P3P and PnP problems are the P3L and PnL problems which are solved for the minimal and overconstrained line correspondence cases respectively. The former is solved by finding the roots to a degree 8 polynomial \cite{P3L}\cite{PnL}, while the latter involves solving higher degree polynomials or large matrix nullspaces \cite{PnL2}\cite{PnL}. Given the similarities between the point and line problems, notably the linear nature of the constraint equations, solving pose problems with both features has also been explored. Zhou \cite{minimal_pose} and Ramalingam \cite{minimal_pose2} both solved point/line combinations for minimal cases via an octic polynomial. Ansar \cite{PnPnL} solved for both points and lines for $n\ge4$ in O($n^2$) time through a matrix nullspace approach, and Vakhitov \cite{PnPnL2} formulated a O($n$) solution by integrating the ends of line segments as points into existing PnP solvers.

With the increasing ubiquity of integrated IMU sensors in devices such as smartphones, drones, and virtual reality systems, measurements of gravity or heading direction have become readily available in many real-world applications. Consequently, incorporating this information as a known prior into pose estimation has attracted growing interest in recent years. Kukelova \cite{Kukelova}, Sweeney \cite{P2P}, and D’Alfonso \cite{P2P3P} provided analytical solutions for the two-point correspondence (P2P) case by solving a simple quadratic equation. Horanyi \cite{gPnLup} addressed the overconstrained scenario for $n \ge 3$ lines using an algebraic approach that reduces the problem to a single cubic polynomial. Lecrosnier \cite{VPnL} also tackled the the same scenario by first estimating the best unconstrained rotation and then projecting it to the nearest valid rotation matrix. Finally, D’Alfonso \cite{IMUPnP} proposed a solution for the PnP problem with axis prior, though it is only locally optimal and requires initial solutions.

We introduce a novel solution to the pose estimation problem with axis prior. Our method accommodates both point and line features in minimal and overconstrained configurations, recovers all finite global solutions, and operates efficiently in $O(n)$ time. The following sections present the mathematical formulation and empirical validation supporting these contributions. Reference code available at \url{https://www.github.com/akschion/PoseGravity}.

\section{Method}
\subsection{Problem Setup} \label{sec:setup}
The linear projective constraints that define the pose estimation problem given \emph{n} point correspondences and \emph{m} line correspondences are typically given by:
\begin{align}
[\mathbf{p}_{i}]_{\times}(\mathbf{R}\mathbf{d}_{i} + \mathbf{T}) &= 0 \label{eq:1}\\
\mathbf{n}_{j}^T (\mathbf{R}\mathbf{m}_{j} + \mathbf{T}) &= 0 \label{eq:2}\\
\mathbf{n}_{j}^T (\mathbf{R}\mathbf{v}_{j}) &= 0 \label{eq:3}\\
\forall i \in 1 \dots n \quad \forall j &\in 1 \dots m \nonumber
\end{align}
where $\mathbf{R}$ and $\mathbf{T}$ are the rotation matrix and translation of the camera's pose to be estimated, $\mathbf{p} = [p_{x}, \ p_{y}, \ p_{z}]^{T}$ and $\mathbf{d} = [d_{x}, \ d_{y}, \ d_{z}]^{T}$ are the given 2D and 3D point correspondences respectively, $\mathbf{n} = [n_{x}, \ n_{y}, \ n_{z}]^{T}$ is a normal vector defining the 2D projective line, and $\mathbf{m} = [m_{x}, \ m_{y}, \ m_{z}]^{T}$ and $\mathbf{v} = [v_{x}, \ v_{y}, \ v_{z}]^{T}$ are the points and line directions defining the 3D lines respectively. The 2D features are assumed to be in normalized coordinates (\ie calibrated camera). $[\mathbf{p}]_{\times}$ denotes the skew-symmetric matrix created from vector $\mathbf{p}$ representing a cross product. It is well known that \cref{eq:1} yields two algebraically independent linear equations for each point correspondence ($[\mathbf{p}]_{\times}$ is a matrix of rank 2), and \cref{eq:2,eq:3} each yield an algebraically independent linear equation for each line correspondence. Therefore, each line or point correspondence gives two pieces of information about the problem.

Given a normalized axis prior $\mathbf{g} = \mathbf{[}g_{x}, \ g_{y}, \ g_{z} \mathbf{]}^{T}$, we can, without loss of generality, assume that this prior measures the world y-axis and is given simply as the camera's y-axis (\ie $\mathbf{g}=[0,1,0]^T$). This is because the 2D and 3D features can be rotated to align with such an intermediate frame, and all resulting solutions can be subsequently rotated back to the original coordinate system. For example, given a general gravity vector measurement $\mathbf{g}$ measuring the world y-axis, we can define a rotation matrix $\mathbf{R_g}$ such that $\mathbf{R_g}\mathbf{g} = [0, 1, 0]^T$. The 2D features $\mathbf{p}_i$ and $\mathbf{n}_j$ are then transformed by $\mathbf{R_g}$, and the estimated pose ($\mathbf{R}, \mathbf{T}$) is subsequently mapped back via $\mathbf{R_g}^T$. See Suppl. Mat. for details on constructing such transformations. With this canonical form, the remaining rotational degree of freedom is about the prior axis, implying that $\mathbf{R}$ takes the form:
\begin{align}
    \mathbf{R} &= \begin{bmatrix} x & 0 & y \\ 0 & 1 & 0 \\ -y & 0 & x \end{bmatrix}\label{eq:6}
\end{align}

Here, \emph{x} and \emph{y} are the cosine and sine of the angle of rotation about the y-axis related elementarily by $x^{2} + y^{2} = 1$.

\subsection{Constrained Least Squares Formulation}
If we have two independent 2D-3D correspondences, \cref{eq:1,eq:2,eq:3} hold exactly. However, if we have more than two, then the left and right sides of those equations may not be equal, and we would seek to estimate the best $\mathbf{R}$ and $\mathbf{T}$ in the least squares sense. Following the formulation of \cite{SQPnP}, we note that a rotation of a vector $\mathbf{u}$ by $\mathbf{R}$ (\ie $\mathbf{Ru}$) can be equivalently written as:
\begin{align}
    \begin{bmatrix}
        u_{x} & u_{z} & 0 \\
        0 & 0 & u_{y} \\
        u_{z} & -u_{x} & 0
    \end{bmatrix}
    \begin{bmatrix}
        x \\ y \\ 1
    \end{bmatrix}
    = \mathbf{U}\mathbf{r} \label{eq:15}
\end{align}
We refer to $\mathbf{U}$ as the matrix representation of vector $\mathbf{u}$. Substituting this form into \cref{eq:1,eq:2,eq:3}, the left-hand sides yield a set of residual error expressions that are linear in $\mathbf{r}$ and $\mathbf{T}$. To obtain the total cost, each expression is multiplied on the left by its transpose, and the results are summed across all point and line correspondences yielding the total sum of squared residuals. This leads to an objective function in the following form:
\begin{gather}
    \min_{\mathbf{r},\mathbf{T}}
    \sum_{i}(\mathbf{D}_{i}\mathbf{r} + \mathbf{T})^{T}\mathbf{Q}_{i}^{p}(\mathbf{D}_{i}\mathbf{r} + \mathbf{T})\space + \nonumber\\
    \sum_{j}(\mathbf{M}_{j}\mathbf{r} + \mathbf{T})^{T}\mathbf{Q}_{j}^{l}(\mathbf{M}_{j}\mathbf{r} + \mathbf{T})\space +
    \delta^{2}\sum_{j}(\mathbf{V}_{j}\mathbf{r})^{T}\mathbf{Q}_{j}^{l}(\mathbf{V}_{j}\mathbf{r}) \label{eq:16}
\end{gather}
where $\mathbf{D}_{i}$, $\mathbf{M}_{j}$, $\mathbf{V}_{l}$ are the matrix representations of $\mathbf{d}_{i}$, $\mathbf{m}_{j}$, and $\mathbf{v}_{j}$ respectively, and $\mathbf{Q}_{i}^{p}$ and $\mathbf{Q}_{j}^{l}$ are $[\mathbf{p}_{i}]_{\times}^{T}[\mathbf{p}_{i}]_{\times}$ and $\mathbf{n}_{j}\mathbf{n}_{j}^{T}$ respectively. $\delta$ is an optional scaling factor (see Suppl. Mat. for details). We observe that this sum of squared linear equations is non-negative and convex with respect to unconstrained $\mathbf{T}$. Taking the derivative of the objective function with respect to $\mathbf{T}$ and setting it to 0, we can solve for the globally minimizing value of $\mathbf{T}$ given $\mathbf{r}$:
\begin{gather}
    \mathbf{S} = -(\sum_{i}\mathbf{Q}_{i}^{p} + \sum_{j}\mathbf{Q}_{j}^{l})^{-1} (\sum_{i}\mathbf{Q}_{i}^{p}\mathbf{D}_{i} + \sum_{j}\mathbf{Q}_{j}^{l}\mathbf{M}_{j}) \label{eq:17a}\\
    \mathbf{T} = \mathbf{S}\mathbf{r} \label{eq:17b}
\end{gather}
If we are given at least two points, three lines, or a point and a line (features assumed linearly independent), then $\sum_{i}\mathbf{Q}_{i}^{p} + \sum_{j}\mathbf{Q}_{j}^{l}$ is guaranteed to be invertible (proof in Suppl. Mat.). Substituting this value of $\mathbf{T}$ back into the objective function and consolidating the terms, we end up with the following constrained quadratic minimization problem:
\begin{gather}
    \mathbf{\Omega} = \sum_{i}(\mathbf{D}_{i} + \mathbf{S})^{T}\mathbf{Q}_{i}^{p}(\mathbf{D}_{i} + \mathbf{S})\space + \nonumber\\
    \sum_{j}(\mathbf{M}_{j} + \mathbf{S})^{T}\mathbf{Q}_{j}^{l}(\mathbf{M}_{j} + \mathbf{S})\space +
    \delta^{2}\sum_{j}\mathbf{V}_{j}^{T}\mathbf{Q}_{j}^{l}\mathbf{V}_{j} \label{eq:18}\\
    \min_{\mathbf{r}} \:
    \mathbf{r}^{T}\mathbf{\Omega}\mathbf{r} \; s.t. \; x^{2} + y^{2} = 1 \label{eq:19}
\end{gather}

\subsection{Solution} \label{sec:solution}
Using the fact that last element of $\mathbf{r}$ is 1, we can rewrite \cref{eq:19} to isolate the variable $\mathbf{x} = [x,y]^T$:
\begin{align}
    \min_{\mathbf{x}} f(\mathbf{x}) = \mathbf{x}^T \mathbf{A}\mathbf{x} + 2\mathbf{b}^T\mathbf{x} + C \;\; s.t. \;\; \mathbf{x}^T\mathbf{x} = 1 \label{eq:matrix_opt}
\end{align}
where $\mathbf{A}$ is the leading $2\times2$ principal submatrix of $\mathbf{\Omega}$, $\mathbf{b}$ is a column vector composed of the last elements of the first two rows of $\mathbf{\Omega}$, and $C$ is the bottom-right entry of $\mathbf{\Omega}$. The problem can be solved via the method of Lagrange multipliers, forming the associated Lagrangian:
\begin{align}
    \mathcal{L}(\mathbf{x}, \lambda) = \mathbf{x}^T \mathbf{A}\mathbf{x} + 2\mathbf{b}^T\mathbf{x} + C - \lambda(\mathbf{x}^T\mathbf{x} - 1) \label{eq:lagrangian}
\end{align}
where $\lambda$ is the Lagrange multiplier. As both the objective and the constraint are differentiable and the constraint gradient ($\nabla_{\mathbf{x}}(\mathbf{x}^T\mathbf{x} - 1) = 2\mathbf{x}$) is nonzero on the feasible set, the first-order optimality condition $\nabla_{\mathbf{x}}\mathcal{L}=0$ yields:
\begin{align}
    (\mathbf{A} - \lambda\mathbf{I})\mathbf{x} = -\mathbf{b}  \label{eq:kkt}
\end{align}
where $\mathbf{I}$ is the $2\times2$ identity matrix. Since \cref{eq:kkt} (along with the feasibility condition $\nabla_{\lambda}\mathcal{L}=\mathbf{x}^T\mathbf{x}-1=0$) provides only the necessary conditions for optimality, the minimum can be identified by comparing objective function values at candidate solutions to other feasible points. Denoting by $\mathbf{x}^*$ any unit-norm point satisfying \cref{eq:kkt}, we substitute the value of $\mathbf{b}$ from \cref{eq:kkt} into \cref{eq:matrix_opt} to obtain:
\begin{gather*}
    f(\mathbf{x}) = \mathbf{x}^{T}\mathbf{A}\mathbf{x} - 2\mathbf{x}^{*T}(\mathbf{A} - \lambda\mathbf{I})\mathbf{x} + C \nonumber \\
    = \mathbf{x}^{T}(\mathbf{A} - \lambda\mathbf{I})\mathbf{x} +\lambda - 2\mathbf{x}^{*T}(\mathbf{A} - \lambda\mathbf{I})\mathbf{x} + C \nonumber \\
    = (\mathbf{x} - \mathbf{x}^*)^T(\mathbf{A} - \lambda\mathbf{I})(\mathbf{x} - \mathbf{x}^*) +\lambda - \mathbf{x}^{*T}(\mathbf{A} - \lambda\mathbf{I})\mathbf{x}^* + C
\end{gather*}
utilizing $\mathbf{x}^T\mathbf{x} = 1$ for feasible points. Similarly:
\begin{gather*}
    f(\mathbf{x}^*) = \mathbf{x}^{*T}\mathbf{A}\mathbf{x}^* - 2\mathbf{x}^{*T}(\mathbf{A} - \lambda\mathbf{I})\mathbf{x}^* + C \nonumber \\
    = \mathbf{x}^{*T}(\mathbf{A} - \lambda\mathbf{I})\mathbf{x}^* + \lambda - 2\mathbf{x}^{*T}(\mathbf{A} - \lambda\mathbf{I})\mathbf{x}^* + C \nonumber\\
    = \lambda - \mathbf{x}^{*T}(\mathbf{A} - \lambda\mathbf{I})\mathbf{x}^* + C \nonumber
\end{gather*}
using $\mathbf{x}^{*T}\mathbf{x}^* = 1$ again for feasible solutions. Subtracting $f(\mathbf{x}^*)$ from $f(\mathbf{x})$ produces the following statement:
\begin{gather}
    f(\mathbf{x}) - f(\mathbf{x}^*) = (\mathbf{x} - \mathbf{x}^*)^T(\mathbf{A} - \lambda\mathbf{I})(\mathbf{x} - \mathbf{x}^*) \label{eq:global_sol}
\end{gather}
Since $\mathbf{\Omega}$ takes the form of a sum of Gram matrices, it is symmetric positive semidefinite. Consequently, its leading principal submatrix $\mathbf{A}$ is also symmetric positive semidefinite. Thus, if there exists a stationary point ($\mathbf{x}^*$, $\lambda$) of \cref{eq:lagrangian} where $\lambda$ is less than or equal to the smallest eigenvalue of $\mathbf{A}$, then from \cref{eq:global_sol} it follows for all feasible points $\mathbf{x}$:
\begin{gather}
    \mathbf{A} - \lambda\mathbf{I} \succeq 0 \implies f(\mathbf{x}) \geq f(\mathbf{x}^*) \label{eq:global_min}
\end{gather}
implying that $\mathbf{x}^*$ is a global minimizer by definition. \par
To further analyze the problem, we transform the objective into a simplified space. As $\mathbf{A}$ is symmetric positive semidefinite, it admits an orthogonal diagonalization $\mathbf{A} = \mathbf{P}^T\mathbf{\Lambda}\mathbf{P}$ where $\mathbf{\Lambda} = diag(\lambda_1, \lambda_2)$ is a diagonal matrix consisting of the eigenvalues of $\mathbf{A}$ ordered such that $\lambda_1 \le \lambda_2$, and $\mathbf{P}$ is an orthogonal matrix whose rows are the corresponding eigenvectors of $\mathbf{A}$. Since $\mathbf{A}$ is only a $2\times2$ matrix, $\mathbf{\Lambda}$ and $\mathbf{P}$ can be computed cheaply in closed form. Performing substitutions $\mathbf{y} = \mathbf{P}\mathbf{x}$ and $\mathbf{c} = \mathbf{P}\mathbf{b}$, we can rewrite the objective and \cref{eq:kkt} as:
\begin{gather}
    f(\mathbf{x}) = g(\mathbf{y}) = \mathbf{y}^T\mathbf{\Lambda}\mathbf{y} + 2\mathbf{c}^T\mathbf{y} + C \label{eq:y_obj}\\
    (\mathbf{\Lambda} - \lambda\mathbf{I})\mathbf{y} = -\mathbf{c} \label{eq:kkt2} \\
    \implies (\lambda_1 - \lambda)y_1 = -c_1, \;\; (\lambda_2 - \lambda)y_2 = -c_2 \label{eq:kkt3}
\end{gather}
Before solving for $\mathbf{y}$, we must determine whether $\lambda$ may be in $spec(\mathbf{A})$. With the simple conditions in \cref{eq:kkt3}, it is apparent that $\lambda$ can only take on a value of $\lambda_1$ or $\lambda_2$ if $c_1=0$ or $c_2=0$ respectively. Thus, we can inspect $\mathbf{c}$ for nonzero elements and proceed to the corresponding case below.

\subsubsection{Case 1: \texorpdfstring{$\lambda \notin spec(\mathbf{A})$}{λ not in spec(A)}} \label{sec:case_1}
In this case, $\mathbf{\Lambda} - \lambda\mathbf{I}$ is nonsingular, so the optimal $\mathbf{y}$ can be written as:
\begin{gather}
   y_1 = \frac{-c_1}{\lambda_1 - \lambda}, \;\; y_2 = \frac{-c_2}{\lambda_2 - \lambda} \label{eq:y_sol}
\end{gather}
For values of $\lambda$ to be feasible, they must produce unit-norm solutions $\mathbf{x}^T\mathbf{x}=1$ (and consequently $\mathbf{y}^T\mathbf{y}=1$). Thus, the problem reduces to finding the roots of the secular equation:
\begin{gather}
    \psi(\lambda) = \frac{c_1^2}{(\lambda_1 - \lambda)^2} + \frac{c_2^2}{(\lambda_2 - \lambda)^2} - 1 \label{eq:secular}
\end{gather}
Clearing the denominators can transform \cref{eq:secular} into a quartic polynomial in $\lambda$ which may have up to four real solutions. To better understand the solution placement, we analyze $\psi(\lambda)$ along with its derivative:
\begin{gather}
    \psi'(\lambda) = \frac{2c_1^2}{(\lambda_1 - \lambda)^3} + \frac{2c_2^2}{(\lambda_2 - \lambda)^3}
\end{gather}
which reveals several key properties for the desired domain ($\lambda < \lambda_1)$: $\psi(\lambda)$ is smooth and strictly increasing ($\psi'(\lambda) > 0$), $\psi(\lambda) \to -1$ as $\lambda \to -\infty$, and $\psi(\lambda) \to \infty$ as $\lambda \to \lambda_1$. Consequently, there exists a unique root $\lambda^*$ of $\psi(\lambda)$ in the open domain (-$\infty$, $\lambda_1$), and it is the smallest root of $\psi(\lambda)$. In principle, this root can be obtained in closed form via the quartic form of $\psi(\lambda)$. However, given our existence and uniqueness guarantees on the restricted domain, iterative approaches such as Newton’s or Halley’s method can more efficiently and reliably obtain $\lambda^*$ from $\psi(\lambda)$ with only a few iterations. Our fast root-finding strategy with guaranteed convergence is detailed in Suppl. Mat. With $\lambda^*$, the associated $\mathbf{x}^*$ can be recovered from \cref{eq:kkt}, which by \cref{eq:global_sol,eq:global_min} is the unique global minimizer since  $\mathbf{A}-\lambda^*\mathbf{I} \succ 0$.

\subsubsection{Case 2: Potentially \texorpdfstring{$\lambda \in spec(\mathbf{A})$}{λ in spec(A)}} \label{sec:case_2}
\textbf{Case 2a $\mathbf{c_1=0, c_2 \neq 0}$}: Substituting these values into \cref{eq:y_obj} and using the constraint $y_1^2 = 1 - y_2^2$, the objective reduces to a univariate quadratic:
\begin{gather}
    g(y_2) = \lambda_1 + (\lambda_2 - \lambda_1)y_2^2 + 2c_2 y_2 + C \label{eq:c1_0}
\end{gather}
In this case, the constrained minimizer of \cref{eq:c1_0} is given by projecting the unconstrained minimizer $\frac{-c_2}{\lambda_2 - \lambda_1}$ to the feasible set $-1\leq y_2 \leq 1$ (proof in Suppl Mat.):
\begin{gather}
    y_2 = \begin{cases}
        \operatorname{min}(\operatorname{max}(\frac{-c_2}{\lambda_2 - \lambda_1}, -1), 1) & \text{if } \lambda_1 \neq \lambda_2 \\
        -\operatorname{sign}(c_2) & \text{if } \lambda_1 = \lambda_2
    \end{cases} \label{eq:sol_c1_0}
\end{gather}
Subsequently, if $|y_2| < 1$, we can get two solutions for $y_1$ as $\pm \sqrt{1 - y_2^2}$ which yields $\lambda=\lambda_1$ from \cref{eq:kkt3}. Otherwise, we obtain a single solution with $y_1=0$, $\lambda = \lambda_2 - |c_2|$. Note that in the latter case, $|c_2| \geq \lambda_2 - \lambda_1$ from \cref{eq:sol_c1_0}, so $\lambda \leq \lambda_1$. \par
\textbf{Case 2b $\mathbf{c_1 \neq 0, c_2 = 0}$}: Analogous to before, we obtain a univariate objective in $y_1$:
\begin{gather}
    g(y_1) = \lambda_2 + (\lambda_1 - \lambda_2)y_1^2 + 2c_1 y_1 + C \label{eq:c2_0}
\end{gather}
However, this time the objective is concave ($g(y_1)'' \leq 0$). For a concave function over a convex set, the minimum is attained at an extreme point of the feasible interval, \ie the boundary $\{-1, 1\}$ (proof in Suppl. Mat.). This results in a single solution, given trivially by $y_1 = -\operatorname{sign}(c_1), y_2=0$ with corresponding multiplier $\lambda = \lambda_1 - |c_1|$ by \cref{eq:kkt3}. \par
\textbf{Case 2c $\mathbf{c_1=0, c_2=0}$}: $||\mathbf{c}|| =0$ implies $||\mathbf{b}|| = 0$. Our original objective simplifies to:
\begin{gather}
    f(\mathbf{x}) = \mathbf{x}^T\mathbf{A}\mathbf{x} + C \label{eq:planar_obj}
\end{gather}
From \cref{eq:kkt}, any stationary point in this case satisfies $\mathbf{A}\mathbf{x}^* = \lambda \mathbf{x}^*$, suggesting that $\mathbf{x}^*$ is an eigenvector of $\mathbf{A}$. Substituting this relation into \cref{eq:planar_obj} shows that the minimum is attained at $\lambda = \lambda_1$, with $\mathbf{x}^*$ being any normalized eigenvector in the associated eigenspace. If $\lambda_1\neq\lambda_2$, the eigenspace of $\lambda_1$ is one-dimensional, and its intersection with the unit circle produces two antipodal solutions, $\mathbf{x}^*$ and $-\mathbf{x}^*$. If $\lambda_1=\lambda_2$, the eigenspace is two-dimensional, coinciding with the entire space of $\mathbf{x}$, so any unit vector is optimal, yielding infinite solutions. \par
Thus, the cases above exhaust all possibilities, and in each one, a solution is obtained with $\lambda \leq \lambda_1$, ensuring we achieve the global minimum by \cref{eq:global_min}. When $\lambda < \lambda_1$, the minimizer is unique, whereas when $\lambda = \lambda_1$, there can be up to two distinct finite solutions (proof in Suppl. Mat.).

\subsection{Special Configurations}
We highlight two physically meaningful special configurations and their connection to the previous solutions.
\subsubsection{Minimal Features}
When the inputs consist of two points ($n = 2$) or one point and one line ($n=1,m=1$), the problem is said to be \emph{minimal} as these features constrain exactly four degrees of freedom— the same number as the problem itself. This configuration commonly arises when only two features are available (\eg baskets on a basketball court, eyes on a face) or when quick estimates are needed for RANSAC iterations or initial solutions of related problems. For minimal configurations, $\mathbf{\Omega}$ from our original objective in \cref{eq:19} has a rank of at most 1 (proof in the Suppl. Mat.). Consequently, the rank of $\mathbf{A}$ is at most 1. Examining the earlier algorithm, we observe that constructing $c_1$ for a singular $\mathbf{A}$ essentially involves computing the determinant of a submatrix of $\mathbf{\Omega}$, which is zero as guaranteed by $\mathbf{\Omega}$’s rank. Thus, this configuration corresponds to Case~2a or~2c from \cref{sec:case_2}. \par
Instead of applying the general algorithm, we can exploit the fact $rank(\mathbf{\Omega})\leq 1$ and express the objective as:
\begin{gather}
    \min_{\mathbf{r}}\: \mathbf{r}^T\mathbf{\Omega}\mathbf{r} = \mathbf{r}^T(\mathbf{q}\mathbf{q}^T)\mathbf{r} = (\mathbf{q}^{T}\mathbf{r})^2 \; s.t. \; x^{2} + y^{2} = 1 \label{eq:31}
\end{gather}
for some vector $\mathbf{q}$. The exact solutions of \cref{eq:31} satisfy $\mathbf{q} \cdot \mathbf{r} = 0$, \ie $\mathbf{q}$ defines a projective line whose intersections with the unit circle (the feasible set) yield solutions to our constrained problem. This leads to a simple quadratic equation that can produce up to two distinct solutions. In practice, $\mathbf{q}$ can be efficiently obtained by selecting any nonzero row of $\mathbf{\Omega}$ since the scale of projective lines is arbitrary. Under certain conditions (\eg infeasible configuration geometry, large noise, \etc), the line may not intersect the circle at all. In such cases, we can still recover the optimal least-squares solution by taking the nearest point on the circle to the line (proof in Suppl. Mat.).

\subsection{Planar Configurations}
Another special case occurs when all of the 3D features lie in the plane $Y=0$ (or equivalently, in any plane orthogonal to the world $Y$-axis since the problem can be translated to such a frame). This arises for instance when detecting features on a ground plane (\eg road markers, field landmarks in football, etc.) while observing an IMU gravity measurement. Inspecting the construction of $\mathbf{\Omega}$ in this scenario, we find that the last columns of $\mathbf{D}_i$, $\mathbf{M}_j$, and $\mathbf{V}_j$ vanish, forcing $||\mathbf{b}|| = 0$. Thus, this situation is an example of Case~2c from earlier, and as such, the solution is the eigevector of $\mathbf{A}$ corresponding to its smallest eigenvalue. Since $\mathbf{A}$ is $2\times2$, this can be expressed simply in closed form as\footnote{Note the solution assumes $a_{1,2} \neq 0$. Otherwise, the solution is either $[1, 0]^T$ or $[0, 1]^T$ depending on which is smaller of $a_{1,1}$ and $a_{2,2}$}:
\begin{gather}
    \mathbf{\Tilde{x}} = \begin{bmatrix} -2a_{1,2} \\ a_{1,1} - a_{2,2} + \sqrt{(a_{1,1} - a_{2,2})^{2} + 4a_{1,2}^2} \end{bmatrix} \label{eq:34}
\end{gather}
where $a_{i,j}$ denotes the element in the $i$th row and $j$th column of $\mathbf{A}$, and $\mathbf{\Tilde{x}}$ is the unnormalized form of $\mathbf{x}^*$.

\begin{figure}[t]
    \centering

    \begin{subfigure}[t]{0.48\columnwidth}
        \centering
        \includegraphics[width=0.95\linewidth]{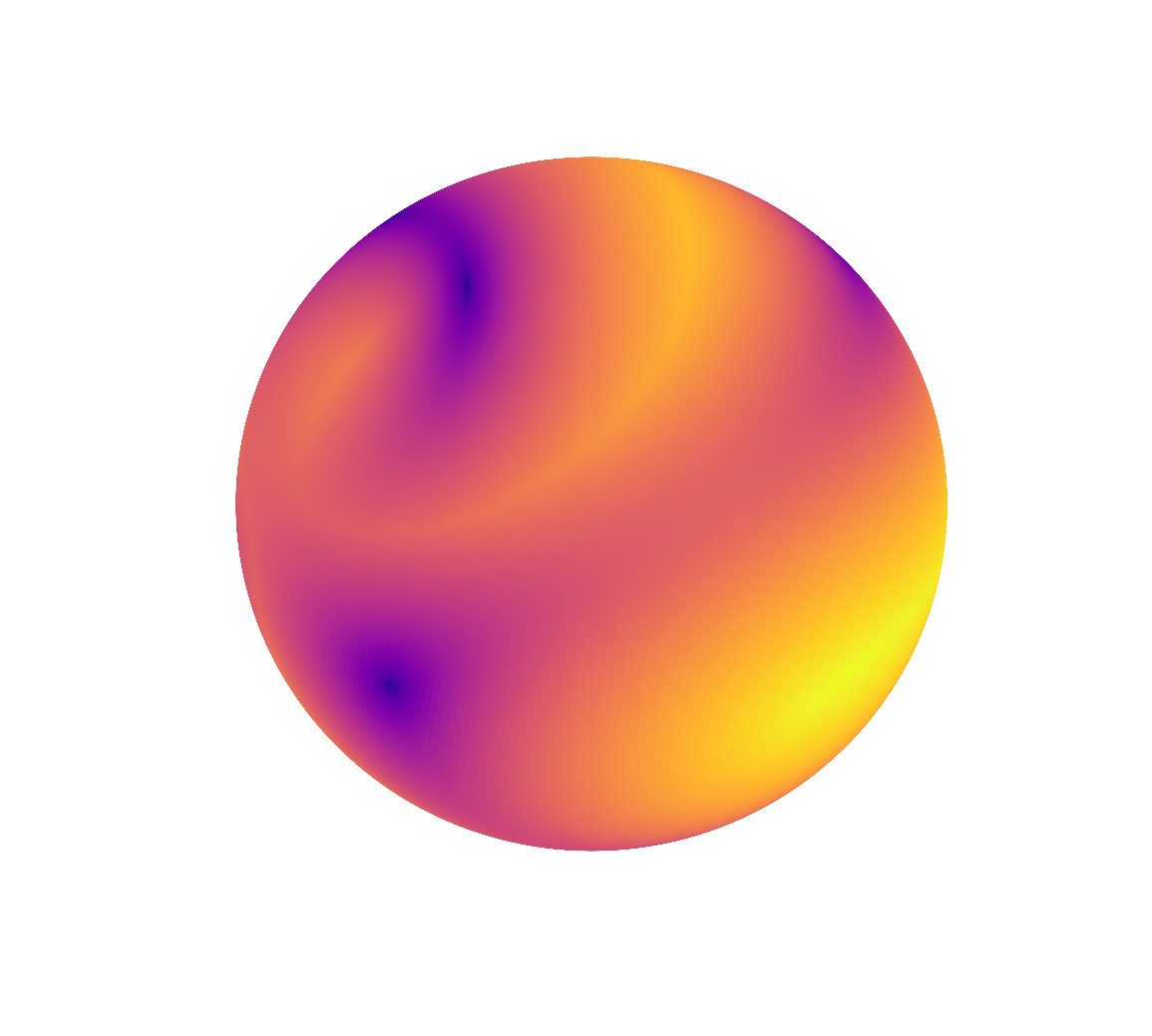}
        \caption{$n=3, m=0$}
    \end{subfigure}\hfill
    \begin{subfigure}[t]{0.48\columnwidth}
        \centering
        \includegraphics[width=0.95\linewidth]{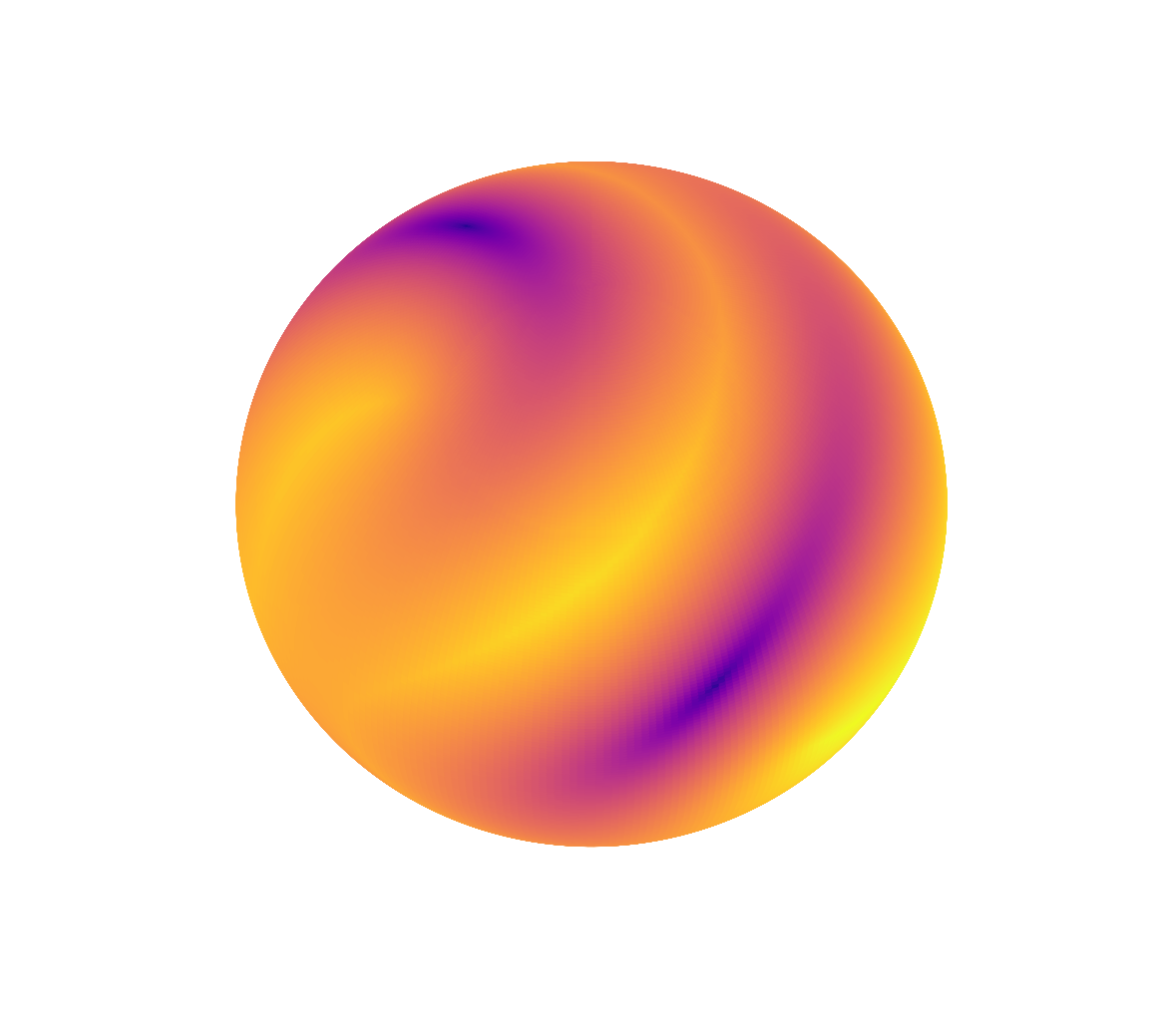}
        \caption{$n=0, m=3$}
    \end{subfigure}

    \vspace{0.1em}

    \begin{subfigure}[t]{0.48\columnwidth}
        \centering
        \includegraphics[width=0.95\linewidth]{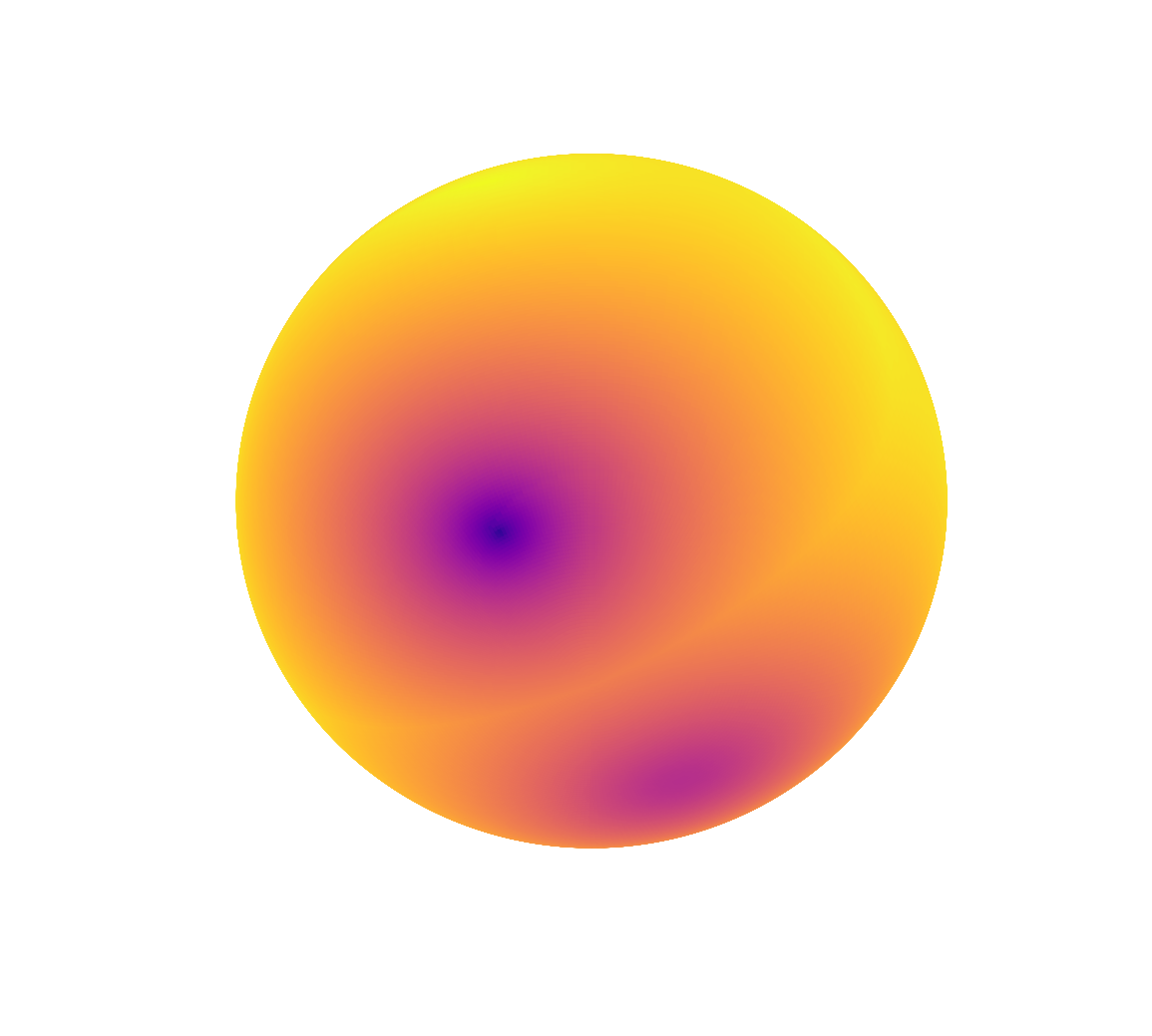}
        \caption{$n=100, m=0$}
    \end{subfigure}\hfill
    \begin{subfigure}[t]{0.48\columnwidth}
        \centering
        \includegraphics[width=0.95\linewidth]{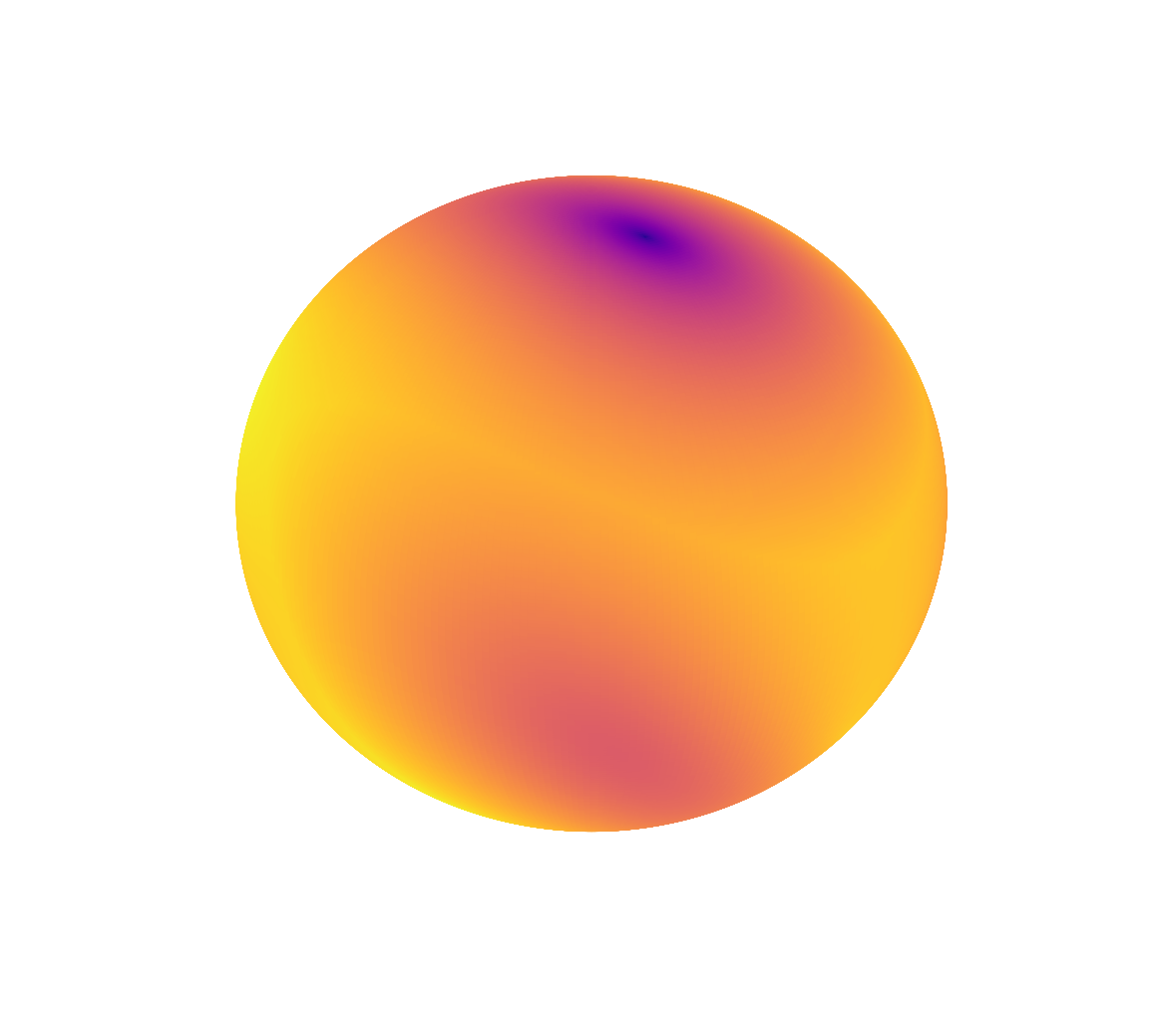}
        \caption{$n=1, m=3$}
    \end{subfigure}

    \caption{Visualization of loss landscapes over all axis priors ($S^2$) for sample problems with $n$ points, $m$ lines. Darker regions correspond to relatively lower objective values. (a)–(b) exhibit multiple strong local minima, (c)–(d) show a single pronounced minimum.}
    \label{fig:PnPnL_loss}
\end{figure}

\section{Relation to General Problem} \label{sec:general_relation}
As noted previously, our objective is derived from the same formulation used in~\cite{SQPnP} for the PnP problem. Thus, we optimize the same loss function as in the general case, but conditioned on an axis prior. This conditioning effectively expresses the solution to the PnPnL problem as a function of the two degrees of freedom defining the axis prior, reducing the problem's dimensionality. A noteworthy benefit of this reduction is the ability to directly visualize the loss landscape (see \cref{fig:PnPnL_loss}). Since the space of axis priors is two-dimensional, the objective values can be readily plotted over a sphere or plane (\eg via stereographic projection), providing intuitive insight into the problem structure. For example, features such as the number of local minima and the sensitivity to axis prior noise can be easily determined by visual inspection for given configurations. \par
Beyond interpretability, the reduction in dimensionality can also offer more efficient optimization strategies for the PnPnL problem. For instance, Newton's method becomes more attractive for pose refinement over Gauss–Newton or Levenberg–Marquardt methods since the Hessian is only $2\times2$. The planar case is particularly appealing for this as the Jacobian and Hessian admit straightforward analytic expressions, and the objective value $\lambda_1 + C$ is simpler for optimization.
\section{Synthetic Experiments}
Synthetic experiments are conducted to evaluate the approach and compare it to existing solutions. Three input types in particular are considered: image plane detections, spherical detections, and planar scene configurations.
\subsection{Experiment Setup}
For each problem, the ground truth rotation is sampled randomly as a quaternion from the unit sphere $S^{3}$, and translations are sampled randomly as a unit vector on $S^{2}$. The ground truth gravity vector measurement is taken as the second column of the rotation matrix (world y-axis). In image plane detection problems, $X$ and $Y$ image point coordinates are uniformly sampled between -1 and 1 with $Z=1$. They are then inversely transformed by the sampled rotation and translation and projected into 3D space by a uniformly random depth between 0.01 and 100. Spherical detection problems are generated similarly but the 2D detections are first sampled uniformly from the sphere $S^{2}$ instead of the image plane to simulate wide-angle cameras. Finally, planar scene problems are generated similarly to the spherical detection case except the transformed points are projected onto the 3D plane $Y=0$. Image points here are checked to be consistent with cheirality constraints. In order to maintain a larger variance in scale, the translation in the planar scene case is scaled by a sampled factor between 0.01 and 100. Line features in 2D and 3D are created from pairs of points treated by the procedure above for each case. \par
To simulate noisy measurements and detections, gravity vector measurement noise and detection noise are added. For the former, the gravity vector measurement is rotated around a random axis by a normally distributed angle with standard deviation $\theta_{noise}$ in degrees. For the latter, independent Gaussian noise with standard deviation $\epsilon_{noise}$ is added to each component of each 2D feature. \par
The accuracy is evaluated by measuring the error in the estimated rotation and translation separately. The rotation error is given as the angle in degrees between the estimated rotation and the ground truth rotation. The translation error is simply the Euclidean distance between the estimated translation and the ground truth translation.
\begin{gather}
    \theta_{err} =  cos^{-1}(\frac{Tr(\mathbf{R}_{gt}^{T}\mathbf{R}_{est}) - 1}{2}) \label{eq:35} \\
    T_{err} = ||\mathbf{T}_{gt} - \mathbf{T}_{est}|| \label{eq:36}
\end{gather}
In the case that more than one solution is returned, the solution with the lowest rotation error is used.

\subsection{Implementation}
We compare the performance (speed and accuracy) of the proposed solver against several existing solutions. Specifically, VPnL \cite{VPnL} and gPnLup \cite{gPnLup} were chosen as comparative solvers accommodating $n\ge 3$ line features and an axis prior. Sweeney \cite{P2P} was chosen as a comparative P2P solution solving minimal point configurations with axis prior. All three were re-implemented and optimized in C++ in a similar style as our algorithm, minimizing compute and avoiding external libraries. For VPnL, the VPnL\textunderscore LS algorithm was used in order to compare analytic solutions, and the SVD operation which was used for estimating nearest rotation was replaced by the FOAM method in \cite{FOAM} which yielded a significant compute improvement for the same accuracy. For gPnLup, the best cubic root is selected by lowest projective error (\ie square of \cref{eq:1,eq:2,eq:3}) akin to the original implementation. gPnLup as presented does not handle multiple solutions from a planar configuration as there is typically a single global minimum for its loss among the roots. Thus, we also return any second best root as well for planar configurations for fairer comparison.\par
Furthermore, Lambda Twist \cite{LambdaTwist} and SQPnP \cite{SQPnP} are also compared as general P3P and PnP solutions respectively to illustrate the utility of axis prior solutions. Production implementations in C++ were taken from the paper authors. Note that SQPnP requires the use of external library Eigen to perform linear algebra operations. \par
All experiments were compiled in C++ 14 with Apple clang 13.0 using the compiler flags {\fontfamily{qcr}\selectfont -O3} and {\fontfamily{qcr}\selectfont -funroll}. For each experiment, 100k trials are run and the median for each metric is reported.

\subsection{Results}

\begin{table}[htb]
    \centering
    \small
    \setlength\tabcolsep{2pt}
    \begin{tabular}{cc|lccc}
        \toprule
        $\epsilon_{noise}$ & $\theta_{noise}$ & Algorithm & \# With Sol. & $\theta_{err}$ & $T_{err}$ \\
        \midrule
        & & Sweeney~\cite{P2P} & 96653 & 1.1263 & 1.28064 \\
        1e{-3} & 1 & Ours* & 96653 & 1.1263 & 1.28064 \\
        & & Ours & 100000 & 1.14786 & 1.32045 \\
        \midrule
        & & Sweeney~\cite{P2P} & 88173 & 12.76779 & 15.24925 \\
        0.1 & 10 & Ours* & 88173 & 12.76779 & 15.24925 \\
        & & Ours & 100000 & 13.57351 & 16.1364 \\
        \bottomrule
    \end{tabular}
    \label{tab:your_label_here}
    \caption{Comparison of two-point solvers on spherical detections with different noise levels. "\# With Sol." reports the number of trials with at least one solution returned. * denotes proposed algorithm without recovery of non-exact solutions.}
    \label{tab:1}
\end{table}

\cref{tab:1} compares the proposed algorithm with a competing P2P method~\cite{P2P} in the minimal two-point case. Our algorithm by default recovers both exact and non-exact solutions, producing a solution in every trial. However, when modified to only return exact solutions, the results of both methods become virtually identical. Since recovered solutions are noisy by definition, their inclusion marginally increases the reported error proportional to their amount in each batch of trials. \par

\begin{table*}[htb]
    \centering
    \small
    \setlength\tabcolsep{2.7pt}
    \begin{tabular}{{@{}cc|c|ccc|ccc@{}}}
        \toprule
        & & & \multicolumn{3}{c|}{$m=3$} & \multicolumn{3}{c}{$m=100$}\\
        $\epsilon_{noise}$ & $\theta_{noise}$ & Config & gPnLup & VPnL\textunderscore LS & Ours & gPnLup & VPnL\textunderscore LS & Ours \\
        \midrule
        & & Image & 0.1012/\underline{0.2696} & \textbf{0.0992}/\textbf{0.2408} & \underline{0.1012}/0.2701  & 0.0891/0.0805 & \textbf{0.0870}/\textbf{0.0528} & \underline{0.0890}/\underline{0.0799} \\
        1e-3 & 0.1 & Spherical & \textbf{0.1009}/0.1917 & 0.1036/\textbf{0.1469} & \underline{0.1011}/\underline{0.1913}  & 0.0875/0.0198 & \underline{0.0875}/\textbf{0.0107} & \textbf{0.0874}/\underline{0.0198} \\
        & & Planar & \underline{0.0953}/\underline{0.1409} & 0.1052/0.1517 & \textbf{0.0953}/\textbf{0.1408}  & 0.087/0.0274 & 0.0922/0.0454 & \textbf{0.087}/\textbf{0.0274} \\
        \midrule
        & & Image & 5.082/25.27 & \textbf{4.148}/\textbf{17.69} & \underline{4.358}/\underline{22.43} & 2.109/4.438 & \textbf{1.032}/\textbf{1.789} & \underline{1.102}/\underline{3.611} \\
        0.01 & 1 & Spherical & 5.115/16.72 & \underline{4.986}/\textbf{10.35} & \textbf{4.382}/\underline{14.88}  & 2.093/1.693 & \underline{1.094}/\textbf{0.8594} & \textbf{1.078}/\underline{1.534} \\
        & & Planar & \underline{3.203}/13.48 & 5.281/\underline{13.35} & \textbf{2.897}/\textbf{12.54} & 1.863/2.363 & \underline{1.139}/\underline{2.013} & \textbf{1.009}/\textbf{1.510} \\
        \midrule
        & & Image & 81.79/90.08 & \underline{77.12}/\textbf{48.32} & \textbf{74.94}/\underline{85.56} & 82.24/49.19 & \underline{29.44}/\textbf{28.44} & \textbf{16.00}/\underline{44.75} \\
        1 & 10 & Spherical & 81.95/77.85 & \textbf{73.02}/\textbf{33.14} & \underline{75.25}/\underline{72.06} & 82.06/10.71 & \underline{22.00}/\underline{10.04} & \textbf{14.85}/\textbf{9.392} \\
        & & Planar & \underline{51.8}/74.65 & 75.05/\textbf{54.69} & \textbf{31.39}/\underline{65.90} & 89.94/42.83 & \underline{25.59}/\underline{30.32} & \textbf{11.58}/\textbf{25.60} \\
        \bottomrule
    \end{tabular}
    \caption{Comparison against other axis prior line-based solvers for different line amounts $m$, noise levels ($\epsilon_{noise}, \theta_{noise}$), and input configurations. Values reported as $\theta_{err}$/$T_{err}$ pairs. Bold font indicates best value, underline indicates next best value among solvers for $\theta_{err}$ and $T_{err}$ separately. gPnLup from \cite{gPnLup}, VPnL\textunderscore LS from \cite{VPnL}.}
    \label{tab:2}
\end{table*}

\cref{tab:2} compares the proposed solver against competing axis-prior line-based methods across a range of experiments. Overall, the solver demonstrates competitive performance, achieving the lowest rotation error in two-thirds of the tests and ranking within the top two in nearly all cases. It performs particularly well in planar configurations where it achieves notably low rotation errors. We attribute this to our specialized solution for the planar case, which otherwise tends to produce poor solutions if not handled carefully. The solver also exhibits improved accuracy as the noise levels and number of lines increase, highlighting its ability to effectively exploit the available geometric information. In contrast, the gPnLup formulation exhibits a bias in its objective with respect to the rotation angle being estimated, leading to weaker performance in most experiments and reduced robustness at higher noise levels. Conversely, VPnL\textunderscore LS performs strongly overall, especially in image-plane scenes and in translation estimation, owing the latter likely to its Plücker line–based formulation which leverages two degrees of freedom per line to estimate translation. However, its performance on rotation estimation can be weaker at times, particularly in planar configurations.

\begin{table}[ht!]
    \centering
    \small
    \setlength\tabcolsep{5pt}
    \begin{subtable}[t]{\linewidth}
        \centering
        \begin{tabular}{{@{}clcc|c@{}}}
            \toprule
            $n$ & Reference solver & Ref. & Ours & Ours (Planar) \\
            \midrule
            2 & P2P~\cite{P2P} & 0.625 & \textbf{0.459} & 0.459 \\
            \midrule[0.25pt]
            3 & Lambda Twist~\cite{LambdaTwist} & 2.083 & \textbf{0.583} & 0.500 \\
            \midrule[0.25pt]
            20 & SQPnP~\cite{SQPnP} & 448.375 & \textbf{1.166} & 1.125 \\
            250 & SQPnP~\cite{SQPnP} & 695.084 & \textbf{9.125} & 9.000 \\
            \bottomrule
        \end{tabular}
        \caption{}
        \label{tab:3a}
    \end{subtable}

    \vspace*{2mm}

    \begin{subtable}[t]{\linewidth}
        \centering
        \begin{tabular}{{@{}c|ccc|c@{}}}
            \toprule
            $m$ & gPnLup \cite{gPnLup} &  VPnL\textunderscore LS \cite{VPnL} & Ours & Ours (Planar) \\
            \midrule
            3 & 0.665 & 0.875 & \textbf{0.584} & 0.541 \\
            20 & 1.791 & 2.375 & \textbf{1.291} & 1.250 \\
            250 & 16.208 & 21.665 & \textbf{10.750} & 10.584 \\
            \bottomrule
        \end{tabular}
        \caption{}
        \label{tab:3b}
    \end{subtable}
    \caption{Median algorithm runtimes in microseconds for a) points and b) lines. (Planar) indicates planar config. used instead of image plane config. For PnP timings, the relevant competing solver is used for each regime of $n$. Bold indicates best (planar excluded).}
\end{table}

\cref{tab:3a,tab:3b} compare runtime timings of each algorithm. For point-based solvers, our algorithm significantly outperforms other solutions which feature more complicated formulations. SQPnP in particular has many expensive operations including a $9\times9$ matrix eigendecomposition. For line-based solvers, the algorithms are all generally fast for small numbers of lines. For larger line amounts, the proposed solver scales more efficiently due to estimating rotation and translation simultaneously with a single pass over the inputs. The other solvers require at least two iterations (three for gPnLup with multiple cubic roots) due to their sequential estimation of rotation followed by translation. Furthermore, the proposed algorithm's streamlined solution for planar configurations, a case ignored by other approaches, yields the best timings in all experiments. However, with our fast root finding strategy, our timings for general configurations follow very closely.

\begin{table*}[t]
    \centering
    \scriptsize
    \setlength\tabcolsep{2.5pt}
    \begin{subtable}[t]{\linewidth}
        \centering
        \resizebox{\linewidth}{!} {
            \begin{tabular}{@{}lc|cccc |cccc |cccc@{}}
                \toprule
                & &
                \multicolumn{4}{c}{gPnLup~\cite{gPnLup}} &
                \multicolumn{4}{c}{VPnL\textunderscore LS~\cite{VPnL}} &
                \multicolumn{4}{c}{Ours} \\
                \cmidrule(lr){3-6} \cmidrule(lr){7-10} \cmidrule(lr){11-14}
                Scene & \# & Loss & $\theta_{err}$ & $T_{err}$ & Time & Loss & $\theta_{err}$ & $T_{err}$ & Time  & Loss & $\theta_{err}$ & $T_{err}$ & Time \\
                \midrule
                Merton College I & 3 &
                6.427e-4 & 0.78242 & \underline{0.07938} & 250.25 &
                \underline{5.764e-4} & \underline{0.18413} & 0.23054 & \underline{78.709} &
                \textbf{5.68e-4} & \textbf{0.17581} & \textbf{0.0303} & \textbf{50.958} \\
                Merton College II & 3 &
                4.83e-4 & 0.80432 & \underline{0.23697} & 247.79 &
                \underline{4.657e-4} & \textbf{0.16426} & 0.44364 & \underline{84.918} &
                \textbf{4.545e-4} & \underline{0.26006} & \textbf{0.05613} & \textbf{53.083} \\
                Merton College III & 3 &
                5.985e-4 & 1.0607 & \underline{0.107} & 157.58 &
                \underline{5.761e-4} & \underline{0.21996} & 0.47652 & \underline{53.958} &
                \textbf{5.725e-4} & \textbf{0.11721} & \textbf{0.03634} & \textbf{30.291} \\
                University Library & 3 &
                \textbf{6.025e-4} & \textbf{0.3041} & \underline{0.08495} & 207.58 &
                1.04e-3 & \underline{0.52204} & 3.7773 & \underline{75.375} &
                \underline{6.17e-4} & 0.56668 & \textbf{0.0639} & \textbf{41.5} \\
                Wadham College & 5 &
                \textbf{1.733e-3} & \textbf{0.18258} & \underline{0.15315} & 387.54 &
                1.846e-3 & 0.96352 & 2.7159 & \underline{155.87} &
                \underline{1.738e-3} & \underline{0.18908} & \textbf{0.14296} & \textbf{89.292} \\
                \bottomrule
            \end{tabular}
        }
        \caption{Axis-prior PnL solvers (lines only). Prior taken from rotation matrix label. Underline indicates second best.}
        \label{tab:vgg_data_lines}
    \end{subtable}

    \vspace*{1mm}

    \begin{subtable}[t]{0.72\linewidth}
        \centering
        \resizebox{\linewidth}{!} {
            \begin{tabular}{@{}lc|cccc |cccc @{}}
                \toprule
                & &
                \multicolumn{4}{c}{SQPnP~\cite{SQPnP}} &
                \multicolumn{4}{c}{Ours w/ Newton} \\
                \cmidrule(lr){3-6} \cmidrule(lr){7-10}
                Scene & \# & Loss & $\theta_{err}$ & $T_{err}$ & Time & Loss & $\theta_{err}$ & $T_{err}$ & Time\\
                \midrule
                Merton College I & 3 &
                3.734e-5 & 0.02348 & 4.241e-4 & 3859 & \textbf{3.732e-5} & \textbf{0.02312} & \textbf{2.874e-4} & \textbf{1297} \\
                Merton College II & 3 &
                4.083e-5 & 0.01922 & 2.174e-3 & 3104 & \textbf{4.077e-5} & \textbf{0.01851} & \textbf{1.952e-3} & \textbf{900}\\
                Merton College III & 3 &
                4.498e-5 & \textbf{0.0} & 1.127e-3 & 3213 & \textbf{4.495e-5} & \textbf{0.0} & \textbf{9.89e-4} & \textbf{1153.5}\\
                University Library & 3 &
                \textbf{7.294e-5} & \textbf{0.0221} & \textbf{6.739e-4} & 3565 & 7.296e-5 & 0.0231 & 1.063e-3 & \textbf{2024} \\
                Wadham College & 5 &
                8.419e-5 & 0.03986 & 1.321e-3 & 6819 & \textbf{8.417e-5} & \textbf{0.03924} & \textbf{1.054e-3} & \textbf{2949}\\
                \bottomrule
            \end{tabular}
        }
        \caption{PnP results (points only). Ours takes $[0, 1, 0]^T$ as prior, then optimizes with Newton's method.}
        \label{tab:vgg_data_points}
    \end{subtable}
    \caption{Updated results on VGG multi-view data. Figures represent cumulative amounts summed across all images for each scene. Loss refers to normalized projective loss, \# refers to number of images in scene, and timings are in microseconds. Bold indicates best.}
    \label{tab:vgg_data}
\end{table*}

\section{Real-World Experiment} \label{sec:real_exp}
Finally, we test our method on the public VGG multi-view dataset~\cite{VGG_MultiViewData}. We select the outdoor scenes (Merton College I-III, University Library, Wadham College), which include 2D/3D point and line correspondences with relatively accurate 3D structure and ground truth camera poses. Our solver is evaluated against other axis-prior line solvers in \cref{tab:vgg_data_lines} and compared to the general point solver in \cref{tab:vgg_data_points}. For the former, the gravity vector is derived from the ground truth rotation. In the latter, we improperly assume that camera aligns with the world y-axis (\ie perfectly upright camera) and subsequently use Newton's method as suggested in \cref{sec:general_relation} to refine the solution. Since the dataset's ``ground truth'' poses and 3D data are derived from a Structure-from-Motion pipeline, we additionally report the normalized projective loss (see Suppl. Mat.). This provides a fairer metric of geometric consistency between the inputs and resulting poses. \par
For line features, our solver achieves the best translation accuracy and overall runtime. It also obtains the lowest projective loss in three of the five scenes and ranks within the top two methods in nearly all remaining cases. The other approaches achieve lower rotation error on some scenes, but this is often at the cost of translation accuracy or runtime. With points, Newton's method converges successfully from our initial estimate in all images and reaches essentially the same minimum as SQPnP. In four of the five scenes, our method attains marginally lower loss and error while being substantially faster overall. This is despite the fact that each Newton iteration requires at least six solves to numerically estimate the Hessian. Furthermore, considering that the initial upright-camera prior was at times up to roughly 12 degrees from the ground truth, these results suggest that the proposed scheme is both efficient and robust.
\section{Conclusion}
This paper introduced a novel pose estimation algorithm which leverages axis prior information such as a gravity vector measurement. Incorporating the axis constraint provides several key benefits over traditional pose estimation methods. These include enabling estimation from fewer feature correspondences, improving robustness to noise, and accelerating iterative pipelines. Extensive experiments validated the algorithm's efficiency and accuracy, especially in planar configurations. A key advantage of this approach is its ability to accommodate arbitrary combinations of point and line features (including points only), which to the authors' knowledge is a first-of-its-kind capability for axis-constrained pose estimation. Furthermore, the identification of minimal cases and planar configurations enables the use of streamlined closed-form solutions, further boosting performance and differentiating the approach. Finally, the algorithm exhibits superior compute efficiency, especially as the number of inputs scales, outperforming comparative approaches and far surpassing general purpose solutions. \par
Potential future research directions include exploring alternative loss formulations to automatically balance the line feature loss terms as well as incorporating prior axis uncertainty into estimation.
{
    \small
    \bibliographystyle{ieeenat_fullname}
    \bibliography{main}
}

 \clearpage
\setcounter{page}{1}
\maketitlesupplementary

\renewcommand{\thesection}{\Alph{section}}
\setcounter{section}{0}

\section{Proofs}
\subsection{Solvability of T}
\textbf{Proposition 1:} \emph{$\sum_{i}\mathbf{Q}_{i}^{p} + \sum_{j}\mathbf{Q}_{j}^{l}$ is invertible for nonzero inputs consisting of at least two independent points, three independent lines, or an independent point and line.} \par

\emph{Proof}
Consider the problem with only $n = 2$ point features $\mathbf{p}_{1}$ and $\mathbf{p}_{2}$ ($\mathbf{p}_{1} \times \mathbf{p}_{2} \neq \mathbf{0}$ by independence). $\mathbf{Q}_{i}^{p} = [\mathbf{p}_{i}]_{\times}^{T}[\mathbf{p}_{i}]_{\times}$ is a symmetric positive semidefinite matrix. Given that $\sum_{i}\mathbf{Q}_{i}^{p}$ is a sum of positive semidefinite matrices:
\begin{gather*}
    ker(\sum_{i}\mathbf{Q}_{i}^{p}) = \bigcap \{ker(\mathbf{Q}_{i}^{p})\} = \bigcap \{\mathbf{p}_{i}\}
\end{gather*}
Two points ($\mathbf{p}_1\times\mathbf{p}_2 \neq 0$ by independence) are sufficient to ensure $dim(ker(\sum_{i}\mathbf{Q}_{i}^{p})) = 0$ which implies $rank(\sum_{i}\mathbf{Q}_{i}^{p}) = 3$ by rank-nullity theorem. \par
Consider the problem with only $m = 3$ line features ($\mathbf{n}_{1} \times \mathbf{n}_{2} \neq \mathbf{0}$, $\mathbf{n}_{2} \times \mathbf{n}_{3} \neq \mathbf{0}$, $\mathbf{n}_{1} \times \mathbf{n}_{3} \neq \mathbf{0}$ by independence). $\sum_{j}\mathbf{Q}_{j}^{l} = \sum_{j}\mathbf{n}_{j}\mathbf{n}_{j}^{T}$ is the sum of real positive semidefinite rank one matrices:
\begin{gather*}
    ker(\sum_{j}\mathbf{Q}_{j}^{l}) = \bigcap \{ker(\mathbf{Q}_{j}^{l})\} = \bigcap \{\mathbf{n}_{j}^{\bot}\} \\
    dim(\bigcap \{\mathbf{n}_{j}^{\bot}\}) = dim(ker(\mathbf{N}))
\end{gather*}
where $\mathbf{N}$ is the $m$ x 3 matrix formed by concatenation of the lines. Thus, three independent lines are sufficient for $rank(\mathbf{N}) = 3$ which implies $dim(ker(\mathbf{N})) = dim(ker(\sum_{j}\mathbf{Q}_{j}^{l})) = 0$ and $rank(\sum_{j}\mathbf{Q}_{j}^{l}) = 3$.\par
Consider the problem with $n = 1$ point $\mathbf{p}_{1}$ and $m = 1$ line $\mathbf{n}_{1}$ ($\mathbf{p}_{1} \cdot \mathbf{n}_{1} \neq 0$ by independence). Using the fact that $\mathbf{Q}_{1}^{p}$ and $\mathbf{Q}_{1}^{l}$ are real positive semidefinite matrices:
\begin{gather*}
    ker(\mathbf{Q}_{1}^{p} + \mathbf{Q}_{2}^{l}) = ker(\mathbf{Q}_{1}^{p}) \cap ker(\mathbf{Q}_{1}^{l}) = \{\mathbf{p}_{1}\} \cap \{ \mathbf{n}_{1}^{\bot}\}
\end{gather*}
Since $\mathbf{p}_{1}$ and $\mathbf{n}_{1}$ are independent, $dim(ker(\mathbf{Q}_{1}^{p} + \mathbf{Q}_{1}^{l})) = 0$, and thus $rank(\mathbf{Q}_{1}^{p} + \mathbf{Q}_{1}^{l}) = 3$. \par
Consider the problem with $n$ points and $m$ lines:
\begin{gather*}
    ker(\sum_{i}\mathbf{Q}_{i}^{p} + \sum_{j}\mathbf{Q}_{j}^{l}) = \Bigl\{\bigcap \{\mathbf{p}_{i}\}\Bigr\} \cap \Bigl\{\bigcap \{\mathbf{n}_{j}^{\bot}\}\Bigr\} \\
    = \mathbf{p}_{1} \cap \dots \cap \mathbf{p}_{n} \cap \mathbf{n}_{1}^{\bot} \cap \dots \cap \mathbf{n}_{m}^{\bot}
\end{gather*}
If there exists a subset of features matching one of the three previous cases, then using the associativity and commutativity of intersection, we can intersect those objects together first creating a trivial kernel $ker(\mathbf{Q})$. Intersecting this kernel with any other kernels would ensure $dim(ker(\sum_{i}\mathbf{Q}_{i}^{p} + \sum_{j}\mathbf{Q}_{j}^{l})) = 0$, extending the solvability to arbitrary supersets of these point and line features.

\subsection{\texorpdfstring{Rank of $\mathbf{\Omega}$}{Rank of Omega} for Minimal Problems} \label{sec:rank_1_proof}
\textbf{Proposition 2:} \emph{For nonzero, independent, and minimal ($n=2,m=0$ or $n=m=1$) inputs, $rank(\mathbf{\Omega}) \le 1$}. \par
\emph{Proof} For the case of two points, using $\mathbf{Q}_{inv} = (\mathbf{Q}_1^p + \mathbf{Q}_2^p)^{-1}$ and $\mathbf{S} = -\mathbf{Q}_{inv}(\mathbf{Q}_1^p\mathbf{D}_1 + \mathbf{Q}_2^p\mathbf{D}_2)$, $\mathbf{\Omega}$ is explicitly expanded to the following terms:
\begin{gather*}
    \mathbf{\Omega} = \sum_i^{2} (\mathbf{D}_i + \mathbf{S})^T\mathbf{Q}_i^p(\mathbf{D}_i + \mathbf{S})\\
    = \mathbf{D}_1^{T}\mathbf{Q}_1^p\mathbf{D}_1 + \mathbf{D}_2^{T}\mathbf{Q}_2^p\mathbf{D}_2 + \mathbf{S}^T(\mathbf{Q}_1^p + \mathbf{Q}_2^p)\mathbf{S} -\\
    2(\mathbf{Q}_1^p\mathbf{D}_1)^T\mathbf{Q}_{inv}\mathbf{Q}_2^p\mathbf{D}_2^{} -2(\mathbf{Q}_2^p\mathbf{D}_2)^T\mathbf{Q}_{inv}\mathbf{Q}_1^p\mathbf{D}_1^{} -\\
    2(\mathbf{Q}_1^p\mathbf{D}_1)^T\mathbf{Q}_{inv}\mathbf{Q}_1^p\mathbf{D}_1^{} -2(\mathbf{Q}_2^p\mathbf{D}_2)^T\mathbf{Q}_{inv}\mathbf{Q}_2^p\mathbf{D}_2^{} \\
    = \mathbf{D}_1^{T}\mathbf{Q}_1^p\mathbf{D}_1 + \mathbf{D}_2^{T}\mathbf{Q}_2^p\mathbf{D}_2 -\\
    (\mathbf{Q}_1^p\mathbf{D}_1)^T\mathbf{Q}_{inv}\mathbf{Q}_2^p\mathbf{D}_2^{} -(\mathbf{Q}_2^p\mathbf{D}_2)^T\mathbf{Q}_{inv}\mathbf{Q}_1^p\mathbf{D}_1^{} -\\
    (\mathbf{Q}_1^p\mathbf{D}_1)^T\mathbf{Q}_{inv}\mathbf{Q}_1^p\mathbf{D}_1^{} -(\mathbf{Q}_2^p\mathbf{D}_2)^T\mathbf{Q}_{inv}\mathbf{Q}_2^p\mathbf{D}_2^{} \\
    = \mathbf{D}_1^{T}(\mathbf{Q}_1^p - \mathbf{Q}_1^{p}\mathbf{Q}_{inv}\mathbf{Q}_1^p)\mathbf{D}_1 - \\\mathbf{D}_2^{T}\mathbf{Q}_2^{p}\mathbf{Q}_{inv}\mathbf{Q}_1^p\mathbf{D}_1^{} - \mathbf{D}_1^{T}\mathbf{Q}_1^{p}\mathbf{Q}_{inv}\mathbf{Q}_2^p\mathbf{D}_2^{} +\\
    \mathbf{D}_2^{T}(\mathbf{Q}_2^p - \mathbf{Q}_2^{p}\mathbf{Q}_{inv}\mathbf{Q}_2^p)\mathbf{D}_2
\end{gather*}
Using $\mathbf{K}_k$ for $k=1\dots4$ to indicate the inner parts between the $\mathbf{D}$ matrices of each term above, we can write:
\begin{gather*}
   \mathbf{\Omega} = \mathbf{D}_1^{T}\mathbf{K}_1\mathbf{D}_1 - \mathbf{D}_2^{T}\mathbf{K}_2\mathbf{D}_1 -\mathbf{D}_1^{T}\mathbf{K}_3\mathbf{D}_2 + \mathbf{D}_2^{T}\mathbf{K}_4\mathbf{D}_2
\end{gather*}
We next examine the $\mathbf{K}_k$ matrices. Taking the difference of the first one with the others yields:
\begin{align*}
    \mathbf{K}_1 - \mathbf{K}_2 &= (\mathbf{Q}_1^p - \mathbf{Q}_1^{p}\mathbf{Q}_{inv}\mathbf{Q}_1^p) - \mathbf{Q}_2^{p}\mathbf{Q}_{inv}\mathbf{Q}_1^p \\
    &= \mathbf{Q}_1^p - (\mathbf{Q}_1^p  + \mathbf{Q}_2^p)\mathbf{Q}_{inv}\mathbf{Q}_1^p = \mathbf{Q}_1^p - \mathbf{Q}_1^p = 0, \\
    \mathbf{K}_1 - \mathbf{K}_3 &= (\mathbf{Q}_1^p - \mathbf{Q}_1^{p}\mathbf{Q}_{inv}\mathbf{Q}_1^p) - \mathbf{Q}_1^{p}\mathbf{Q}_{inv}\mathbf{Q}_2^p \\
    &= \mathbf{Q}_1^p - \mathbf{Q}_1^p\mathbf{Q}_{inv}(\mathbf{Q}_1^p  + \mathbf{Q}_2^p) = \mathbf{Q}_1^p - \mathbf{Q}_1^p = 0, \\
    \mathbf{K}_1 - \mathbf{K}_4 &= \mathbf{K}_1 - (\mathbf{Q}_2^p - \mathbf{Q}_2^{p}\mathbf{Q}_{inv}\mathbf{Q}_2^p) \\
    &= \mathbf{K}_1 - \mathbf{Q}_2^p + \mathbf{Q}_2^{p}\mathbf{Q}_{inv}(\mathbf{Q}_1^p + \mathbf{Q}_2^p) - \mathbf{K}_2 \\
    &= \mathbf{K}_1 -  \mathbf{Q}_2^p +  \mathbf{Q}_2^p - \mathbf{K}_2 = 0
\end{align*}
Thus, all the inner matrices are equivalent ($\mathbf{K}_k =\mathbf{K}$). Furthermore, we can examine $\mathbf{K}_2$ or $\mathbf{K}_3$ to see that the two image points $\mathbf{p}_1, \mathbf{p}_2 \in ker(\mathbf{K})$ trivially. Since the points are assumed independent (\ie $\mathbf{p}_1 \times \mathbf{p}_2 \neq 0$), $dim(ker(\mathbf{K})) \ge 2$, so the rank of $\mathbf{K}$ is at most 1. We can now rewrite the loss as:
\begin{gather*}
    \mathbf{\Omega} = (\mathbf{D}_1^{} - \mathbf{D}_2^{})^T\mathbf{K}(\mathbf{D}_1^{} - \mathbf{D}_2^{})
\end{gather*}
Since $\mathbf{K}$ has at most rank 1, and multiplying by other matrices cannot increase the rank, $rank(\mathbf{\Omega}) \le 1$. \par
For the case of one point and one line, denoting $\mathbf{Q}_{inv}=(\mathbf{Q}_1^p + \mathbf{Q}_1^l)^{-1}$ and $\mathbf{S} = -\mathbf{Q}_{inv}(\mathbf{Q}_1^p\mathbf{D}_1 + \mathbf{Q}_1^l\mathbf{M}_1)$, the explicit terms of $\mathbf{\Omega}$ are:
\begin{gather*}
    \mathbf{\Omega} = (\mathbf{D}_1 + \mathbf{S})^T\mathbf{Q}_1^p(\mathbf{D}_1 + \mathbf{S}) + \\ (\mathbf{M}_1 + \mathbf{S})^T\mathbf{Q}_1^l(\mathbf{M}_1 + \mathbf{S}) + \delta^2\mathbf{V}_1^T\mathbf{Q}_1^l\mathbf{V}_1 \\
    = \mathbf{D}_1^{T}\mathbf{Q}_1^p\mathbf{D}_1 +  \mathbf{M}_1^{T}\mathbf{Q}_1^l\mathbf{M}_1+\delta^2\mathbf{V}_1^T\mathbf{Q}_1^l\mathbf{V}_1  + \\\mathbf{S}^T\mathbf{Q}_1^p\mathbf{S} + \mathbf{S}^T\mathbf{Q}_1^l\mathbf{S} -\\
    2(\mathbf{Q}_1^p\mathbf{D}_1)^T\mathbf{Q}_{inv}\mathbf{Q}_1^p\mathbf{D}_1^{} - 2(\mathbf{Q}_1^l\mathbf{M}_1)^T\mathbf{Q}_{inv}\mathbf{Q}_1^p\mathbf{D}_1^{} - \\2(\mathbf{Q}_1^p\mathbf{D}_1)^T\mathbf{Q}_{inv}\mathbf{Q}_1^l\mathbf{M}_1^{} - 2(\mathbf{Q}_1^l\mathbf{M}_1)^T\mathbf{Q}_{inv}\mathbf{Q}_1^l\mathbf{M}_1^{} \\
    = \mathbf{D}_1^{T}\mathbf{Q}_1^p\mathbf{D}_1 +  \mathbf{M}_1^{T}\mathbf{Q}_1^l\mathbf{M}_1+\delta^2\mathbf{V}_1^T\mathbf{Q}_1^l\mathbf{V}_1 - \\
    (\mathbf{Q}_1^p\mathbf{D}_1)^T\mathbf{Q}_{inv}\mathbf{Q}_1^p\mathbf{D}_1^{} - (\mathbf{Q}_1^l\mathbf{M}_1)^T\mathbf{Q}_{inv}\mathbf{Q}_1^p\mathbf{D}_1^{} - \\(\mathbf{Q}_1^p\mathbf{D}_1)^T\mathbf{Q}_{inv}\mathbf{Q}_1^l\mathbf{M}_1^{} - (\mathbf{Q}_1^l\mathbf{M}_1)^T\mathbf{Q}_{inv}\mathbf{Q}_1^l\mathbf{M}_1^{} \\
    =\mathbf{D}_1^{T}(\mathbf{Q}_1^p - \mathbf{Q}_1^{p}\mathbf{Q}_{inv}\mathbf{Q}_1^p)\mathbf{D}_1 - \\\mathbf{M}_1^{T}\mathbf{Q}_1^{l}\mathbf{Q}_{inv}\mathbf{Q}_1^p\mathbf{D}_1^{} - \mathbf{D}_1^{T}\mathbf{Q}_1^{p}\mathbf{Q}_{inv}\mathbf{Q}_1^l\mathbf{M}_1^{} +\\
    \mathbf{M}_1^{T}(\mathbf{Q}_1^l - \mathbf{Q}_1^{l}\mathbf{Q}_{inv}\mathbf{Q}_1^l)\mathbf{M}_1 + \delta^2\mathbf{V}_1^T\mathbf{Q}_1^l\mathbf{V}_1
\end{gather*}
Denoting $\mathbf{L}_k$ for $k=1\dots4$ to represent the inner parts between the $\mathbf{D}$ and $\mathbf{M}$ matrices in each term above, we can write:
\begin{gather*}
    \mathbf{\Omega}= \mathbf{D}_1^{T}\mathbf{L}_1\mathbf{D}_1 - \mathbf{M}_1^{T}\mathbf{L}_2\mathbf{D}_1 -\mathbf{D}_1^{T}\mathbf{L}_3\mathbf{M}_1 + \\ \mathbf{M}_1^{T}\mathbf{L}_4\mathbf{M}_1 + \delta^2\mathbf{V}_1^T\mathbf{Q}_1^l\mathbf{V}_1
\end{gather*}
Performing the same differences between $\mathbf{L}_1$ and the other $\mathbf{L}_k$ matrices as in the two point case, we arrive at the analogous result that all the $\mathbf{L}_k$ matrices are equivalent ($\mathbf{L}_k=\mathbf{L}$). Examining $\mathbf{L}_2$ or $\mathbf{L}_3$, we see that the image point and the image line's orthogonal complement $\mathbf{p}_1, \mathbf{n}_1^{\bot} \in ker(\mathbf{L})$ trivially and are one and two dimensional respectively. Since the image objects are assumed independent (\ie $\mathbf{p}_1 \cdot \mathbf{n}_1 \neq 0$), the point together with the line's orthogonal complement form a three dimensional basis for the nullspace of $\mathbf{L}$ implying that $\mathbf{L}$ is the zero matrix. As a result, the only remaining term in the expression is the direction loss $\delta^2\mathbf{V}_1^T\mathbf{Q}_1^l\mathbf{V}_1$. Since $\mathbf{Q}_1^l$ has rank 1, and multiplying by other matrices and scaling cannot increase the rank, $rank(\mathbf{\Omega}) \le 1$.
\par
Intuitively, this result stems from the fact that minimal configurations only constrains four degrees of freedom. Estimation of $\mathbf{T}$ consumes three of those constraints, so there is only one left when constructing the 3x3 matrix $\mathbf{\Omega}$. The problem is solved by applying additional constraints to the solution $\mathbf{r}$, specifying that $x^{2} + y^{2} = 1$ and the third element of $\mathbf{r}$ is 1. It is noteworthy that when $rank(\mathbf{K}) = 1$ in the two-point case, the eigenvector associated with the nonzero eigenvalue of $\mathbf{K}$ is given by $\mathbf{p}_{1} \times \mathbf{p}_{2}$. This vector defines the projective line passing through the two points analogous to $\mathbf{n}_1$. Furthermore, $\mathbf{D}_1 - \mathbf{D}_2$ defines the direction of the line passing through the 3D points analogous to $\mathbf{V}_1$. Thus, $\mathbf{\Omega}$ has the same geometric interpretation relating to line features in either case.

\subsection{Proof of Case 2 Solutions}
\subsubsection{Case 2a}
Starting from \cref{eq:c1_0}, we note that:
\begin{gather*}
    g'(y_2) = 2(\lambda_2 - \lambda_1)y_2 + 2c_2 \\
    g''(y_2) = 2(\lambda_2 - \lambda_1)
\end{gather*}
Assume $\lambda_1\neq \lambda_2$. We see that $y_2^* = \frac{-c_2}{\lambda_2 - \lambda_1}$ is the unique stationary point from $g'(y_2^*)=0$, and since $\lambda_1 <\lambda_2$, $g''(y_2)>0$, making the stationary point a minimum. Substituting $c_2 = -(\lambda_2 -\lambda_1)y_2^*$ into \cref{eq:c1_0}, we get:
\begin{gather*}
    g(y_2) = \lambda_1 + (\lambda_2 - \lambda_1)y_2^2 - 2(\lambda_2 -\lambda_1)y_2^* y_2 + C \\
    = \lambda_1 + (\lambda_2 - \lambda_1)(y_2^2 - 2y_2^* y_2) + C \\
    = \lambda_1 + (\lambda_2 - \lambda_1)((y_2 - y_2^*)^2  -y_2^{*2}) + C
\end{gather*}
The only non-constant term in the expression is $(\lambda_2 - \lambda_1)(y_2 - y_2^*)^2$. Thus, the loss is minimized by being as close to $y_2^*$ in Euclidean distance as possible. Since $\mathbf{y}^T\mathbf{y}=1$, $-1\leq y_2 \leq 1$. If $y_2^*$ is in the feasible set, then that is our minimizer. If it is outside the feasible set, $|y_2^*| > 1$, so the closest feasible point has a magnitude of 1 and sign in the direction of $y_2^*$, \ie $-sign(c_2)$ (as the denominator of $y_2^*$ is positive). \par
For $\lambda_1=\lambda_2$, the objective function reduces to $g(y_2) = 2c_2y_2 + \lambda_1+ C$. Thus, the objective is linear in $y_2$, so the minimum value is the largest feasible magnitude value in the direction opposite of $c_2$, \ie $-sign(c_2)$. Combining both parts, we obtain \cref{eq:sol_c1_0}.

\subsubsection{Case 2b}
From \cref{eq:c2_0}, we obtain the derivatives as:
\begin{gather*}
    g'(y_1) = -2(\lambda_2 - \lambda_1)y_1 + 2c_1 \\
    g''(y_1) = -2(\lambda_2 - \lambda_1)
\end{gather*}
Since $\lambda_1\leq \lambda_2$ always, $g''(y_1) \leq 0$ making $g(y_1)$ a concave function. As $\mathbf{y}^T\mathbf{y}=1$, the feasible set is once again $-1\leq y_1 \leq 1$. A property of concave functions is that:
\begin{gather*}
    g((1-\alpha)(\gamma_1) + \alpha(\gamma_2)) \geq (1-\alpha)g(\gamma_1) + \alpha g(\gamma_2)
\end{gather*}
for any two feasible points $\gamma_1,\gamma_2 \in [-1, 1]$ and all $\alpha \in [0,1]$. Taking $\gamma_1,\gamma_2$ to be the boundary points $\{-1, 1\}$:
\begin{gather*}
    g((1-\alpha)(-1) + \alpha(1)) \geq (1-\alpha)g(-1) + \alpha g(1)
\end{gather*}
We note that $(1-\alpha)(-1) + \alpha(1)$ exactly parameterizes the entire feasible set. Taking $m = min(g(-1), g(1))$, we see that:
\begin{gather*}
    (1-\alpha)g(-1) + \alpha g(1) \geq (1 - \alpha)m + \alpha m = m
\end{gather*}
Thus, combining the inequalities, the function over the feasible set is always greater than or equal to minimum value at the boundary. Substituting both boundary values back into \cref{eq:c2_0} and comparing, the only difference is the value of the linear term $2c_1y_1$, so the minimum point is the boundary value with opposite sign of $c_1$, \ie $y_1=-sign(c_1)$.

\subsection{Maximum Number of Finite Global Solutions}
\textbf{Proposition 3:} \emph{The objective in \cref{eq:matrix_opt} has at most two distinct global minimizers; otherwise it has infinitely many.} \par
\emph{Proof} From \cref{eq:global_sol}, it is clear that if $\lambda<\lambda_1$, the global minimizer must be unique as its objective is strictly smaller than that of all other feasible points. Furthermore, all of our cases from \cref{sec:solution} yielded solutions with $\lambda\leq\lambda_1$. Hence, it suffices to consider $\lambda=\lambda_1$. \par
For contradiction, assume that there are at least three but finitely many distinct global minimizers. We denote three of them as $\mathbf{x}_1, \mathbf{x}_2,\mathbf{x}_3$ with $\mathbf{x}_i^T\mathbf{x}_i=1$. Since they have the same objective value, $f(\mathbf{x}_i) - f(\mathbf{x}_j) = (\mathbf{x}_i- \mathbf{x}_j)^T(\mathbf{A} - \lambda\mathbf{I})(\mathbf{x}_i-\mathbf{x}_j) = 0$ for any $i\neq j$. Therefore, $\mathbf{x}_1 - \mathbf{x}_2$ and $\mathbf{x}_1 - \mathbf{x}_3$ both lie in the kernel of $(\mathbf{A}-\lambda\mathbf{I})$. If $\mathbf{x}_1, \mathbf{x}_2,\mathbf{x}_3$ were collinear, their common line would intersect the unit circle in at most two points, contradicting the feasibility or distinctness assumptions. Therefore, $\mathbf{x}_1 - \mathbf{x}_2$ and $\mathbf{x}_1 - \mathbf{x}_3$ are linearly independent, making $dim(ker(\mathbf{A}-\lambda\mathbf{I}))=2$ and $rank(\mathbf{A}-\lambda\mathbf{I})=0$ by rank-nullity, implying that $\mathbf{A}-\lambda\mathbf{I}$ is the zero matrix. However, from \cref{eq:global_sol}, this would make every distinct feasible point $\mathbf{x}$ have the same objective value as $\mathbf{x}_1$, violating the assumption of finite global minimizers. Thus, the problem supports at most two distinct global minimizers or infinitely many.

\subsection{Recovery of Non-Exact Minimal Solutions}
\textbf{Proposition 4:} \emph {For minimal problems ($n=2,m=0$ or $n=m=1$), if there are no exact solutions, then the best feasible solution is the point on the unit circle closest to the intersection of the nullspace of $\mathbf{\Omega}$ and the x{\text -}y plane.} \par
\emph{Proof} \cref{sec:rank_1_proof} proves that $\mathbf{\Omega}$ has at most rank 1 in this case, so we can express $\mathbf{\Omega}=\mathbf{q}\mathbf{q}^T$ for some vector $\mathbf{q}=[q_1, q_2,q_3]^T$. The nullspace of $\mathbf{\Omega}$ is $\mathbf{q}^{\bot}$ whose intersection with the $x{\text -}y$ plane is the projective line $q_1x + q_2y + q_3=0$. The distance between a projective point $\mathbf{r}=[x, y, 1]^T$ and that projective line in the $x{\text -}y$ plane is given as:
\begin{gather*}
    dist_{\mathbf{q}}(\mathbf{r}) = \frac{|q_1x + q_2y + q_3|}{\sqrt{q_1^{2} + q_2^{2}}} \\
    \implies dist_{\mathbf{q}}(\mathbf{r})^2 = \frac{(\mathbf{q}^{T}\mathbf{r})^2}{q_1^{2} + q_2^{2}} = \frac{\mathbf{r}^T\mathbf{\Omega}\mathbf{r}}{q_1^{2} + q_2^{2}}
\end{gather*}
Assuming $q_1^2+q_2^2 \neq 0$, the squared distance is simply the original loss function scaled by a constant. Therefore, minimizing the distance (and thereby distance squared) to the projective line in the $x{\text -}y$ plane minimizes the loss function. As there are no exact solutions, the line does not intersect the unit circle, so the point on the circle closest to the line is the loss-minimizing solution. This solution can be simply calculated as:
\begin{gather*}
    \mathbf{r}^* = \begin{bmatrix} \frac{-q_1}{\sqrt{q_1^{2} + q_2^{2}}} && \frac{-q_2}{\sqrt{q_1^{2} + q_2^{2}}} && 1\end{bmatrix} ^{T}
\end{gather*}
If $q_1 = q_2 = 0$, $\mathbf{q}^{\bot}$ does not intersect the $x{\text -}y$ plane. In this case, $\mathbf{q}^T\mathbf{r}$ is constant, yielding infinitely many solutions.

\section{Additional Implementation Details}
\subsection{Root-Finding Strategy}
In \cref{sec:case_1}, it was shown that for $\lambda\neq \lambda_1$ (which is verified from whether $c_1 \neq 0$), the secular equation (\cref{eq:secular}) is guaranteed to have a unique root $\lambda^* < \lambda_1$. In order to compute this root efficiently, we employ a change of variables $t = \lambda_1 - \lambda > 0$ and $\Delta = \lambda_2 - \lambda_1 \geq 0$ to simplify $\psi(\lambda)$ into:
\begin{gather}
    \phi(t) = \frac{c_1^2}{t^2} + \frac{c_2^2}{(t + \Delta)^2} - 1 \label{eq:secular_t} \\
    \phi'(t) = -\frac{2c_1^2}{t^3} - \frac{2c_2^2}{(t + \Delta)^3} < 0
\end{gather}
From this, we infer that $\phi(t)$ is monotically decreasing, $\phi(t) \to \infty$ as $t \to 0$, and $\phi(t) \to -1$ as $t \to \infty$. Thus, our task is now mapped to finding the unique root $t^*$ of \cref{eq:secular_t} on the domain $t>0$. \par
We establish the following simple lower bounds on $t^*$:
\begin{gather}
    1 = \frac{c_1^2}{t^{*2}} + \frac{c_2^2}{(t^* + \Delta)^2} \geq \frac{c_1^2}{t^{*2}} \nonumber \\
    \implies t^* \geq |c_1| \label{eq:t_lb1}\\
    1 = \frac{c_1^2}{t^{*2}} + \frac{c_2^2}{(t^* + \Delta)^2}  \geq \frac{c_1^2 + c_2^2}{(t^* + \Delta)^2}  \geq \frac{c_2^2}{(t^* + \Delta)^2} \nonumber\\
    \implies t^* \geq |c_2| - \Delta \label{eq:t_lb2}
\end{gather}
making $t^* \geq max(|c_1|, |c_2| - \Delta)$. This bound can serve as a rough initial point for our iteration. \par
Next, we prove that Newton's method is guaranteed to converge when starting from any intial value $0 < t_0 < t^*$. The update step is given by:
\begin{align*}
    \quad t_{n+1} &= t_n - \frac{\phi(t_n)}{\phi'(t_n)}
\end{align*}
Because $\phi'(t)<0$ and $\phi(t_n)>0$ for $t_n<t^*$, the update ensures $t_{n+1} > t_n$. Since $\phi''(t_n) > 0$, the function is strictly convex on this domain, so its tangent line (given by the first-order Taylor approximation about $t_n$) lies strictly below the function for $t\neq t_n$:
\begin{gather*}
    \phi(t) > \phi(t_n) + \phi'(t_n)(t - t_n)
\end{gather*}
Evaluating at $t^*$:
\begin{gather*}
    0 = \phi(t^*) > \phi(t_n) + \phi'(t_n)(t^* - t_n) \\
    \implies t^* > t_n - \frac{\phi(t_n)}{\phi'(t_n)} = t_{n+1}
\end{gather*}
Note the inequality was reversed because $\phi'(t_n) < 0$. Thus, each update of Newton's method is bounded above by $t^*$, establishing $t_n < t_{n+1} < t^*$. Since the sequence $\{t_n\}$ monotonically increases and is bounded above, it converges. By continuity of $\phi(t)$, the only possible limit is $t^*$. Given this result, Newton's method is guaranteed to converge starting from a lower bound on our domain. \par
To refine the initial guess $t_0$, we further analyze the variables defining $\phi$. Empirically, we observe that $|c_1| \gg |c_2|$ in many cases, so the bound given by $|c_1|$ is typically dominant. In this regime, we introduce the parameter $\nu \ge 0$ and define the bivariate function:
\begin{gather*}
    J(t, \nu) = \frac{c_1^2}{t^2} + \frac{\nu}{(t + \Delta)^2} - 1
\end{gather*}
with $\phi(t) = J(t, c_2^2)$ and the root $t^*$ characterized by $J(t^*, c_2^2)=0$. We view $J$ over the domain $t \in (0, \infty), \nu \in [0, \infty)$. Its partial derivatives are given by:
\begin{gather*}
    \frac{\partial J}{\partial t} = -\frac{2c_1^2}{t^3} - \frac{2\nu}{(t + \Delta)^3} < 0, \quad \frac{\partial J}{\partial \nu} = \frac{1}{(t + \Delta)^2} > 0
\end{gather*}
verifying $J(t, \nu)$ is continuous and smooth (\ie $C^1$) over the domain. For each fixed $\nu\geq0$, we see that $J(t, \nu) \to \infty$ as $t\to 0$, $J(t, \nu) \to -1$ as $t \to \infty$, and $\frac{\partial J(t, \nu)}{\partial t} < 0$. Therefore, each $\nu\geq 0$ admits a unique root $t(\nu)$ such that $J(t(\nu), \nu)=0$. In particular, $t(0) = |c_1|$ and $t(c_2^2)= t^*$. We are interested in the behavior of $t(\nu)$ for small nonnegative $\nu$, particularly its one-sided derivative at $\nu=0^+$. As $J(t(\nu),\nu)=0$ for all $\nu \ge 0$, the total derivative with respect to $\nu$ over $\nu > 0$ is 0, yielding the relation:
\begin{gather*}
    \frac{d J(t(\nu), \nu)}{d \nu} = \frac{\partial J(t(\nu), \nu)}{\partial t}\frac{d t(\nu)}{d \nu} + \frac{\partial J(t(\nu), \nu)}{\partial \nu} = 0 \\
    \implies \frac{d t(\nu)}{d \nu} = -\frac{\frac{\partial J(t, \nu)}{\partial \nu}}{\frac{\partial J(t, \nu)}{\partial t}} = \frac{(t + \Delta)^{-2}}{\frac{2c_1^2}{t^3} + \frac{2\nu}{(t + \Delta)^3}}
\end{gather*}
The denominator $\frac{\partial J}{\partial t}$ is strictly nonzero on our domain, so the above expression is well-defined. Furthermore, since $J$ is continuous and has a unique root $t(\nu)$ for each $\nu$, $t(\nu)\to t(0)=|c_1|$ as $\nu\to 0^+$. Therefore, the one-sided derivative $t'(0) = \lim_{\nu\to 0^+} \frac{dt(\nu)}{d\nu}$ exists and is obtained by taking the limit as $(t,\nu)\to(|c_1|,0)$:
\begin{gather*}
    t'(0^+) = \frac{|c_1|}{2(|c_1| + \Delta)^2}.
\end{gather*}
For small $\nu \ge 0$, a first-order Taylor approximation of $t(\nu)$ about $\nu=0$ therefore gives:
\begin{gather}
    t(\nu) \approx t(0) + \frac{\partial t}{\partial \nu}(0^+)(\nu - 0) \nonumber \\
    t^* = t(c_2^2) \approx |c_1| + \frac{|c_1|c_2^2}{2(|c_1| + \Delta)^2} \label{eq:t_ift}
\end{gather}
This gives a more refined initial estimate $t_0 \geq |c_1|$, but it no longer serves as a lower bound on $t^*$. If $t_n > t^*$, then $\phi(t_n)<0$, and the Newton update will yield $t_{n+1} < t_n$, moving towards the root. In this case though, $t_{n+1}$ is not bounded below by $t^*$. If it does not monotonically converge, it overshoots the root and in theory, can land in a region outside of our convergence domain (\eg $t_{n+1} \leq 0$). Although this is unlikely in practice, we can always safeguard the update (\eg $t_{n+1} = max(t_{n+1}, |c_1|)$) such that any overshoot will strictly remain in the region where convergence is guaranteed, ensuring we find the root regardless. \par
In the case that $|c_1|$ is not large relative to $|c_2|$ or $\Delta$ (heuristically tested by whether $|c_1| < |c_2| - \Delta$), we can derive a simple improved bound for $t_0$. Consider the convex function $h(x) = \frac{1}{x^2}$ on the domain $x \in (0, \infty)$ and variables $w_1 = \frac{c_1^2}{c_1^2 + c_2^2}, w_2 = \frac{c_2^2}{c_1^2 + c_2^2}$ so that $w_1 + w_2 = 1$. Applying Jensen's inequality for convex functions at points $x_1=t^*$ and $x_2=t^* + \Delta$ for $\Delta\geq 0$ gives:
\begin{gather}
    h(w_1 t^* + w_2(t^* + \Delta)) \leq w_1 h(t^*) + w_2 h(t^* + \Delta) \nonumber\\
    \frac{1}{(w_1 t^* + w_2t^* + w_2\Delta)^2} \leq  \frac{w_1}{t^{*2}} + \frac{w_2}{(t^* + \Delta)^2} \nonumber \\
    \frac{c_1^2 + c_2^2}{(t^* + w_2\Delta)^2} \leq \frac{c_1^2}{t^{*2}} + \frac{c_2^2}{(t^* + \Delta)^2} = 1 \nonumber \\
    \implies t^* \geq \sqrt{c_1^2 + c_2^2} - w_2\Delta \geq |c_2| -w_2\Delta \label{eq:t_jensen}
\end{gather}
Since $|c_2| - \Delta > |c_1|$ in this case, $t_0$ from \cref{eq:t_jensen} clearly gives a tighter lower bound on $t^*$ than before, and starting from it once again guarantees monotonic convergence. \par
In summary, we have the following strategy for computing our initial estimate:
\begin{gather}
    t_0 = \begin{cases}
              |c_1| + \frac{|c_1|c_2^2}{2(|c_1| + \Delta)^2}  & \text{if } |c_1| \geq |c_2| - \Delta, \\
              |c_2| -\frac{c_2^2}{c_1^2 + c_2^2}\Delta & \text{otherwise}
    \end{cases}
\end{gather}
In both cases, we derive a high-quality initial guess $t_0$ with minimal computational cost, relying solely on basic arithmetic and avoiding square root operations. With this initialization, Newton's method typically converges ($|\phi(t_n)|<10^{-12}$) within five steps. However, despite the theoretical guarantees of Newton's method, we instead employ Halley's method with the same $t_0$ strategy for our experiments as it demonstrates robust and even faster convergence in practice.

\subsection{Transformation to Canonical Frame}
In \cref{sec:setup}, we introduced an intermediate coordinate frame as a way to isolate the remaining rotational degree of freedom. Given a normalized axis prior $\mathbf{g}$ in the camera frame corresponding to a normalized vector $\mathbf{w}$ in the world frame, we can always transform the problem to this frame via rotations aligning both vectors to the y-axis ($\mathbf{e}_y = [0, 1, 0]^T$).
We construct rotation matrices $\mathbf{R_g}$ and $\mathbf{R_w}$ such that $\mathbf{R_g} \mathbf{g} = \mathbf{R_w}\mathbf{w} = \mathbf{e}_y$. Since the rotation aligning two vectors is not unique, we select the \emph{rotation of least angle}. For a generalized unit vector $\mathbf{u}$ (representing $\mathbf{g}$ or $\mathbf{w}$), this rotation's axis $\mathbf{k}$ and angle $\theta$ can be derived from the dot and cross products respectively:
\begin{gather}
    \cos \theta = \mathbf{u} \cdot \mathbf{e}_y, \quad \sin \theta = \|\mathbf{u} \times \mathbf{e}_y\| \\
    \mathbf{k} = \frac{\mathbf{u} \times \mathbf{e}_y}{\|\mathbf{u} \times \mathbf{e}_y\|}
\end{gather}
Substituting these terms into the Rodrigues' rotation formula and simplifying yields a computationally efficient form that involves only basic arithmetic operations:
\begin{equation}
    \mathbf{R_u} = \begin{bmatrix}
                       \frac{u_z^2}{1 + u_y} + u_y & -u_x & \frac{-u_x u_z}{1 + u_y} \\
                       u_x & u_y & u_z \\
                       \frac{-u_x u_z}{1 + u_y} & -u_z & \frac{u_x^2}{1 + u_y} + u_y
    \end{bmatrix} \in SO(3) \label{eq:R_u}
\end{equation}
In the singularity case where $u_y=-1$, we choose $\mathbf{R_u}$ to be a 180 degree rotation about the z-axis (\ie $\mathbf{R_u} = diag(-1, -1, 1)$). \par
Using $\mathbf{R_g}$ and $\mathbf{R_w}$ computed via \cref{eq:R_u}, we define the transformed problem parameters:
\begin{gather}
    \mathbf{\tilde{R}} = \mathbf{R_g}\mathbf{R}\mathbf{R_w}^T,  \quad \mathbf{\tilde{T}} = \mathbf{R_g}\mathbf{T} \label{eq:param_trans} \\
    \mathbf{\tilde{p}}_i = \mathbf{R_g}\mathbf{p}_i,  \quad \mathbf{\tilde{n}}_j = \mathbf{R_g}\mathbf{n}_j \\
    \mathbf{\tilde{d}}_i = \mathbf{R_w}\mathbf{d}_i,  \quad \mathbf{\tilde{m}}_j = \mathbf{R_w}\mathbf{m}_j, \quad \mathbf{\tilde{v}}_j = \mathbf{R_w}\mathbf{v}_j
\end{gather}
Because the rotation $\mathbf{R}$ in our axis-constrained problem aligns $\mathbf{w}$ to $\mathbf{g}$ and both vectors are aligned to the y-axis in our intermediate frame, the new relative rotation $\mathbf{\tilde{R}}$ must be restricted to rotations about the y-axis, assuming the form of \cref{eq:6}.
We justify this coordinate transformation by observing that our objective function is rotationally invariant. For example, considering the point correspondence constraint in \cref{eq:1} with our transformed variables and substituting their definitions from above, we obtain:
\begin{gather*}
    [\mathbf{\Tilde{p}}_i]_{\times}(\mathbf{\Tilde{R}}\mathbf{\Tilde{d}}_i + \mathbf{\Tilde{T}}) = 0 \\
    [\mathbf{R_g} \mathbf{p}_i]_{\times} ((\mathbf{R_g}\mathbf{R}\mathbf{R_w}^T) (\mathbf{R_w} \mathbf{d}_i) + \mathbf{R_g}\mathbf{T}) = 0 \\
(\mathbf{R_g}[\mathbf{p}_i]_{\times}\mathbf{R_g}^T)\bigl( \mathbf{R_g}(\mathbf{R} \mathbf{d}_i + \mathbf{T})\bigr) = 0 \\
    \mathbf{R_g}[\mathbf{p}_i]_{\times}(\mathbf{R} \mathbf{d}_i + \mathbf{T}) = 0
\end{gather*}
which uses the identity $[\mathbf{R}\mathbf{a}]_{\times} =\mathbf{R} [\mathbf{a}]_{\times}\mathbf{R}^T$. Crucially, $\mathbf{R_g}$ is an orthogonal matrix, so it preserves the Euclidean norm ($||\mathbf{R_g}\mathbf{x}|| = ||\mathbf{x}||$). Therefore, the least-squares objective in \cref{eq:16}, constructed via the sum of squared residual norms, is identical in both frames, with the result holding similarly for line constraints as well. Adding the fact that rotational transformations are bijective, we can thus simply solve for the optimal $\mathbf{\Tilde{R}}, \mathbf{\Tilde{T}}$ in the canonical frame and uniquely map the solution back to an optimal pose in the original coordinate systems via $\mathbf{R} = \mathbf{R_g}^T \mathbf{\tilde{R}} \mathbf{R_w}$ and $\mathbf{T} = \mathbf{R_g}^T \mathbf{\tilde{T}}$.

\subsection{Normalized Projective Loss} \label{sec:normalized_proj_loss}
For the real-world experiments, we report a normalized projective loss in addition to rotation and translation errors. The per-image loss used in evaluation is:
\begin{gather}
    \mathbf{\Gamma}(\mathbf{z}) \coloneqq \frac{\mathbf{R}\mathbf{z} + \mathbf{T}}{||\mathbf{R}\mathbf{z} + \mathbf{T}||} \nonumber \\
    \mathcal{L}_{points}(\mathbf{R}, \mathbf{T}) = \sum_i \left\|\mathbf{p}_i \times \mathbf{\Gamma}(\mathbf{d}_i)\right\|^2 \\
    \mathcal{L}_{lines}(\mathbf{R}, \mathbf{T}) = \sum_j \Bigl((\mathbf{n}_j^T\mathbf{R}\mathbf{v}_j)^2  + \left(\mathbf{n}_j^T\mathbf{\Gamma}(\mathbf{m}_j)\right)^2\Bigr) \label{eq:normalized_projective_loss}
\end{gather}
By normalizing the transformed detections before computing the loss, we can measure the geometric alignment without the bias of the scene's position and scale.

\subsection{Note on Line Loss Scale $\delta^2$}
For line features, the proposed solver's loss function combines the rotation and translation loss terms instead of estimating them sequentially like other approaches. Since those two terms in loss function in \cref{eq:16} have different 3D objects associated with them, there can be a bias from the points $\mathbf{m}_j$ and the line directions $\mathbf{v}_j$ having different scales (\eg if $\mathbf{v}_{j}$ is normalized), leading to the loss being dominated by one term and resulting in less accurate pose estimates. Estimates from planar configurations are particularly sensitive to this scale difference. This issue does not occur with point features as their loss is represented by a single term which encapsulates both the rotation and translation components in a naturally balanced way. To compensate for this bias, we found it sufficient to introduce a constant scaling factor $\delta^2$ which manually balances the line direction loss. This parameter may be tuned depending on the specific use case. We use $\delta^2 = 100$ for synthetic experiments and $\delta^2 = 1$ for real-world experiments, both with normalized $\mathbf{v}_j$.

\end{document}